\documentclass[fleqn,10pt]{olplainarticle}
\usepackage{bm}
\usepackage{amsmath}
\usepackage{amssymb}
\usepackage{hyperref}
\usepackage{algorithmic}
\usepackage{algorithm}

\newtheorem{thm}{Theorem}
\newtheorem{lem}{Lemma}
\newtheorem{definition}{Definition}

\newtheorem{prop}{Proposition}
\newtheorem{remark}{Remark}

\usepackage{graphicx}
\usepackage{float} 
\usepackage{subcaption}
\usepackage{multirow}
\usepackage{booktabs}
\usepackage{pifont}
\usepackage[flushleft]{threeparttable}  
\newcommand{\cmark}{\ding{51}}
\newcommand{\xmark}{\ding{55}}

\title{Network Topology Inference from Smooth Signals under Partial Observability}

\author[1]{Chuansen Peng}
\author[1]{Hanning Tang}
\author[1]{Zhiguo Wang}
\author[1]{Xiaojing Shen}
\affil[1]{School of Mathematics, Sichuan University, Chengdu, Sichuan, 610064, China}

\keywords{Graph learning, partial observability, network topology inference}

\begin{abstract}
Inferring network topology from smooth signals is a significant problem in data science and engineering. A common challenge in real-world scenarios is the availability of only partially observed nodes. While some studies have considered hidden nodes and proposed various optimization frameworks, existing methods often lack the practical efficiency needed for large-scale networks or fail to provide theoretical convergence guarantees. In this paper, we address the problem of inferring network topologies from smooth signals with partially observed nodes. We propose a first-order algorithmic framework that includes two variants: one based on column sparsity regularization and the other on a low-rank constraint. We establish $(i)$ linear convergence rates governed by observability and network‐identifiability conditions, and $(ii)$ finite‐sample recovery bounds quantifying how hidden nodes and noise jointly affect reconstruction accuracy. Extensive experiments on both synthetic and real-world data show that our results align with theoretical predictions, exhibiting not only linear convergence but also superior speed compared to existing methods. To the best of our knowledge, this is the first work to propose a first-order algorithmic framework for inferring network structures from smooth signals under partial observability, offering both guaranteed linear convergence and practical effectiveness for large-scale networks. 
\end{abstract}

\begin{document}

\flushbottom
\maketitle
\thispagestyle{empty}

\section{Introduction}\label{sect1}
Network topology inference from smooth signals has emerged as a pivotal area of research in the fields of machine learning and data science \cite{ortega2018graph}, \cite{dong2020graph}, \cite{fortunato2018science}, \cite{10448349}, \cite{araghi2023outlier}. The ability to infer the underlying graph structure from observed signals has profound implications for a wide array of applications \cite{liu2020web}, \cite{liu2019shifu2}, ranging from social network analysis to sensor network localization \cite{shuman2013emerging}, \cite{zhang2024direct}. This task is particularly challenging due to the inherent complexity of graph structures and the variability of signals \cite{liu2016sensor}, \cite{zhang2025graph}.

The primary goal of network topology inference is to uncover the topology that best explains the observed data, often characterized by the smoothness of signals on the graph. This notion of smoothness is typically quantified by the graph Laplacian, which encodes the connectivity and edge weights of the graph \cite{yuan2023joint}. The objective is to learn a graph that minimizes the total variation of the signals, subject to constraints that reflect prior knowledge or desirable properties of the graph \cite{qiu2017time}, \cite{egilmez2017graph}.

Recent advancements in network topology inference have focused on developing efficient and robust algorithms that can handle large-scale datasets and accommodate various types of signals \cite{zhu2022multi}, \cite{kalofolias2016learn}, \cite{6409473}, \cite{6494675}, \cite{li2018resilient}. These methods often leverage optimization techniques \cite{ma2016alternating}, \cite{wang2023linearly}, \cite{wang2021efficient}, spectral graph theory \cite{segarra2017network}, and statistical inference \cite{kolaczyk2014statistical}, \cite{chen2022unbiased}, \cite{talukdar2020physics}, \cite{veedu2023topology}, \cite{li2023topology} to iteratively refine the graph structure based on the observed data. The integration of these diverse mathematical tools has propelled the field forward, enabling the analysis of increasingly complex and high-dimensional datasets \cite{karanikolas2016multi}, \cite{shen2017kernel}, \cite{yu2021distributed}.

Despite notable advancements, several challenges persist in graph learning from smooth signals. A primary concern is the delicate balance between fidelity to observed data and the structural complexity of the inferred graph \cite{chen2020graph}, \cite{su2022comprehensive}. Additionally, the presence of noise and the non-uniqueness of solutions pose significant barriers to obtaining accurate and reliable graph estimates \cite{asif2021graph}, \cite{wu2020comprehensive}. Moreover, the scalability of graph learning algorithms to massive datasets \cite{geisler2021robustness}, \cite{zhou2020graph} and the integration of dynamic \cite{fang2025joint} or multi-modal signals demand sustained research efforts.

Specifically, the standard network-inference approach in existing works often assumes observations from all nodes of the graph are available. However, in certain environments, only observations from a subset of nodes are accessible, with the rest being unobserved or hidden. The existence of hidden nodes introduces a relevant and challenging problem, as closely related signal values from two observed nodes may be influenced not only by each other but also by a third latent node connected to both \cite{buciulea2022learning}. Furthermore, the absence of observations from hidden nodes complicates the network inference problem, rendering it more challenging and ill-posed. Network-inference studies that have addressed the issue of partial observability include examples in the context of Gaussian graphical model selection \cite{chandrasekaran2010latent}, \cite{hentschel2024statistical} or nonlinear regression \cite{mei2018silvar}. 

\begin{table}[!t]
\centering
\caption{Comparison of Different Algorithms for Graph Learning from Smooth Signals}
\label{tab:example_table}
\resizebox{\textwidth}{!}{%
\begin{tabular}{lcccc}
\toprule
\textbf{Algorithm} & \textbf{Partial Observability} & \textbf{First-order Methods} & \textbf{Convergence Rate} & \textbf{Computational Complexity} \\
\midrule
PDGL \cite{kalofolias2016learn} & \xmark  & \cmark & \xmark & $\mathcal{O}(o^2)$\\
GL-SigRep \cite{dong2016learning} & \xmark   & \cmark & \xmark & $\mathcal{O}(o^3)$ \\
pADMM-GL \cite{wang2023linearly} & \xmark & \cmark & \cmark & $\mathcal{O}(o^2)$\\
Gsm \cite{buciulea2022learning}    & \cmark & \xmark & \xmark & $\mathcal{O}(o^7)$ \\
JH-GSR\cite{navarro2024joint}  & \cmark & \xmark & \xmark & $\mathcal{O}(o^7)$ \\
\midrule
\textbf{GLOPSS (Ours)}            & \cmark & \cmark & \cmark & $\mathcal{O}(o^2)$ \\
\bottomrule
\end{tabular}
}
\end{table}

A commonly adopted strategy for solving network topology inference from smooth signals under partial observability involves formulating an optimization problem and then employing a CVX solver to compute solutions \cite{buciulea2022learning}, \cite{10507157}. In practice, we observe that CVX for solving graph learning with hidden variables tends to converge slowly for large-scaled networks. Consequently, there is an urgent imperative to devise algorithms that exhibit rapid convergence, thereby enhancing their applicability and efficiency in real-world scenarios. This recognition underscores the need for methodological innovation in graph learning, aiming to alleviate the computational burden and enable the effective analysis of large-scale graph structures.

In this article, based on the prior work and progress of \cite{buciulea2022learning}, we introduce a first-order optimization framework for inferring network topology from smooth signals under partial observability and provide a rigorous convergence analysis of the algorithms. We summarize our principal contributions as follows:
\begin{itemize}
    \item[$1)$] In response to the practical practical network topology inference problem from smooth signals under partial observability in large-scale networks, we show that the problem can be reformulated as a optimization problem of separable objective functions with multiple blocks and linear equality constraints. Based on this, we propose a first-order algorithmic framework for graph learning from smooth signals under partial observabilty (GLOPSS). In particular, based on different regularization conditions, we present two variants of our algorithm: one is based on column sparsity regularization (GLOPSS-CS), and the other is underpinned by a low-rank constraint (GLOPSS-LR).
    \item[$2)$] Building upon the convergence analysis work in  \cite{wang2023linearly}, \cite{hong2017linear}, we prove the linear convergence rate of GLOPSS. Furthermore, based on the structure  of the optimization problem of GLOPSS, we establish a lower bound on the decrease in the norm of the difference between successive iterations and the optimal solution, and then proving that as long as the step sizes are bounded by the reciprocal of the largest singular value of the coefficient matrices, GLOPSS can linearly converge to the optimal solution of the target problem from any initial point. To the best of our knowledge, as shown in Table \ref{tab:example_table}, GLOPSS is the first provably linearly convergent first-order method for network topology inference from smooth signals under partial observabilty.
    \item[$3)$] We establish rigorous theoretical convergence guarantees with two key contributions. First, we prove that GLOPSS converges linearly at an explicit rate determined by the observability coefficient and network identifiability conditions, thereby characterizing how partial observability modulates convergence speed. Second, we derive finite-sample recovery bounds that explicitly quantify the impact of hidden nodes on reconstruction accuracy under noisy observations, showing that the estimation error scales proportionally with the ratio of hidden to observed nodes.
    \item[$4)$] We provide extensive experiments on both synthetic and real-world data showing that GLOPSS aligns with theoretical predictions, exhibiting not only linear convergence but also superior speed compared to the state-of-the-art methods. The experimental data encompasses scenarios with different network sizes, number of latent nodes, noise power of the measurement. Both GLOPSS-CS and GLOPSS-LR exhibit higher accuracy and robustness, especially in the presence of noise and hidden nodes.
\end{itemize}

These contributions not only advance the theoretical foundations of graph learning but also offer a practical and effective algorithmic solution, underscoring the methodological innovation and potential impact of our approach.

\textbf{\textit{Synopsis:}} The remainder of this paper is organized as follows. In Section \ref{sect2}, we formally introduce the network topology inference problem and present the underlying assumptions, notations and objective functions that underpin our approach. Section \ref{sect3} details the proposed optimization framework: we reformulate the inference task as a separable multi‐block problem with linear constraints and derive a tailored first‐order algorithm, highlighting its computational steps and implementation nuances. In Section \ref{sect4}, we rigorously analyze the convergence properties of the algorithm, proving global convergence under standard assumptions and characterizing the linear convergence rate, and then we derive explicit linear convergence rates and finite-sample recovery bounds quantifying hidden-node effects on accuracy. Section \ref{sect5} presents extensive numerical experiments on both synthetic and real‐world datasets, demonstrating the effectiveness, robustness, and scalability of our method compared to existing approaches. Finally, Section \ref{sect6} concludes the paper, summarizes our main findings, and outlines directions for future research.

\section{Problem Formulation}\label{sect2}
Prior to delineating the problem statement we aspire to resolve, we commence by elucidating certain notations and fundamental concepts.

\subsection{Notations}
For a positive integer $m$, let $[m]:=\{1, 2, \ldots, m\}$ denote the set of integers from $1$ to $m$. The symbol $\mathbf{1}$ (resp. $\mathbf{0}$) represents the all-ones (resp. all-zeros) vector, with its dimensionality determined by the context.

Consider a graph $\mathcal{G}=(\mathcal{V}, \mathcal{E})$, where $\mathcal{V}=[m]$ denotes the set of nodes, and $\mathcal{E}\subseteq\mathcal{V}\times\mathcal{V}$ represents the set of edges. We assume that the graph does not contain self-loops, meaning $(i,i)\notin\mathcal{E}$ for all $i\in\mathcal{V}$. In an undirected graph, relationships among the $m$ vertices are captured by a symmetric, non-negative matrix $\mathbf{W}\in\mathbb{R}^{m\times m}$, where a strictly positive entry $W_{i,j}>0$ indicates the presence of an edge $(i,j)$ in the edge set $\mathcal{E}$. A graph signal is modeled as a vector $\mathbf{x}\in\mathbb{R}^m$, whose $i$-th entry $x_i$ denotes the value observed at node $i$. In many practical scenarios, the true weighting matrix $\mathbf{W}$ is not directly available; thus, the central problem lies in reconstructing this latent connectivity pattern from a dataset of $n$ observed graph signals $\{\mathbf{x}_1, \ldots, \mathbf{x}_n\}\subset\mathbb{R}^m$. The graph Laplacian matrix is defined as $\mathbf{L}:=\mathrm{Diag}(\mathbf{W1})-\mathbf{W}$. We denote the set of combinatorial Laplacians by $\mathcal{L}$, which is defined as:
\begin{align*}
    \mathcal{L}:=\{\mathbf{L}\in\mathbb{R}^{m\times m}\mid L_{ij}\leq0, \text{for}\quad i\neq j; \mathbf{L}=\mathbf{L}^T;\mathbf{L1}=\mathbf{0}; \mathbf{L}\succeq 0\}.
\end{align*}
where $\mathbf{L}\succeq 0$ indicates that the Laplaican matrix is positive semi-definite (PSD).

For vectors $\mathbf{p}$ and $\mathbf{q}$, we use the following notations:
\begin{itemize}
    \item $p_i$ denotes the $i$-th component of $\mathbf{p}$,
    \item $\Vert \mathbf{p}\Vert_2$ represents the $l_2$-norm of $\mathbf{p}$,
    \item $\log(\mathbf{p})$ is element-wise logarithm of $\mathbf{p}$,
    \item $\mathbf{p}^2$ denotes the element-wise square of $\mathbf{p}$,
    \item $\sqrt{\mathbf{p}}$ represents the element-wise square root of $\mathbf{p}$,
    \item  $\mathrm{Diag}(\mathbf{p})$ denotes a diagonal matrix with $\mathbf{p}$ on its diagonal,
    \item $\mathbf{p}/\mathbf{q}$ denotes element-wise division of $\mathbf{p}$ by $\mathbf{q}$,
    \item $\Vert\mathbf{p}\Vert_{\mathbf{Q}}:=(\mathbf{p}^T\mathbf{Q}\mathbf{p})^{1/2}$ denotes the $\mathbf{Q}$-norm of $\mathbf{p}$.
\end{itemize}

For matrices $\mathbf{P}$ and $\mathbf{Q}$, we use:
\begin{itemize}
    \item $P_{ij}$ for the $(i,j)$-th element of $\mathbf{P}$,
    \item $\Vert\mathbf{P}\Vert_F$ for the Forbenius norm of $\mathbf{P}$,
    \item $\Vert\mathbf{P}\Vert_{1,1}$ for the $L_{1,1}$-norm of $\mathbf{P}$,
    \item $\Vert\mathbf{P}\Vert_{2,1}$ for the $L_{2,1}$-norm of $\mathbf{P}$,
    \item $\Vert\mathbf{P}\Vert_2$ for spectral norm of $\mathbf{P}$,
    \item $\Vert\mathbf{P}\Vert_\ast$ for the nuclear norm of $\mathbf{P}$,
    \item $\Vert\mathbf{P}\Vert_{F, off}$ for the Forbenius norm excluding the element of the diagonal of $\mathbf{P}$,
    \item $\text{diag}(\mathbf{P})$ for the vector of diagonal entries of $\mathbf{P}$,
    \item $\mathbf{P}\odot\mathbf{Q}$ for the Hadamard product of $\mathbf{P}$ and $\mathbf{Q}$,
    \item $[\mathbf{P};\mathbf{Q}]:=\begin{bmatrix}
    \mathbf{P}\\
    \mathbf{Q}
\end{bmatrix}$ for the block column matrix formed by $\mathbf{P}$ and $\mathbf{Q}$. 
\end{itemize}

Let $\mathbf{p}$ be a vector in $\mathbb{R}^m$, let $\mathcal{X}\subseteq\mathbb{R}^m$ be any subset of that space, and let $\mathbf{Q}$ be a  symmetric, positive semi-definite matrix in $\mathbb{R}^{m\times m}$. We introduce the indicator function of $\mathcal{X}$ by
\begin{equation*}
    \mathbf{p}\rightarrow\mathbb{I}_{\mathcal{X}}(\mathbf{p}):=\begin{cases}
0,& \text{$\mathbf{p}\in\mathcal{X}$}\\
+\infty,& \text{$\mathbf{p}\notin\mathcal{X}$}
\end{cases}
\end{equation*}
and we measure the proximity of $\mathbf{p}$ to the set $\mathcal{X}$ via
\begin{align*}
    \mathrm{dist}(\mathbf{p},\mathcal{X}):&=\inf\{\Vert\mathbf{p}-\mathbf{q}\Vert_2\mid \mathbf{q}\in\mathcal{X}\},\\
    \mathrm{dist}_{\mathbf{Q}}(\mathbf{p},\mathcal{X}):&=\inf\{\Vert\mathbf{p}-\mathbf{q}\Vert_{\mathbf{Q}}\mid \mathbf{q}\in\mathcal{X}\},
\end{align*}
where $\Vert\mathbf{p}\Vert_{\mathbf{Q}}=\sqrt{\mathbf{p}^T\mathbf{Qp}}$.

For any extended-real-valued function $f:\mathbb{R}^m\rightarrow(-\infty,+\infty]$ that is proper, we denote:
\begin{itemize}
    \item $\mathrm{dom}(f):=\{\mathbf{p}\mid f(\mathbf{p})<+\infty\}$ as its domain,
    \item $\mathrm{prox}_f(\mathbf{p}):=\mathrm{arg}\min_\mathbf{q}\left\{f(\mathbf{q})+\frac{1}{2}\Vert\mathbf{q}-\mathbf{p}\Vert_2^2\right\}$ as the unique minimizer of its the proximal operator at $\mathbf{p}\in \mathrm{dom}(f)$,
    \item $
    \partial f(\mathbf{p}):= \{\mathbf{a}\mid f(\mathbf{q})\geq f(\mathbf{p})+\mathbf{a}^T(\mathbf{q}-\mathbf{p})\quad\text{for all}\quad \mathbf{q}\}$ as the sub-differential of $f$ at $\mathbf{p}\in\mathrm{dom}(f)$.
\end{itemize}

Finally, let $\text{vec}:\mathbb{R}^{m\times m}\rightarrow\mathbb{R}^{m^2}$ denote the usual colum-stacking operator, so that if $\mathbf{W}$ is a $m\times m$ matrix then $\mathbf{w}=\text{vec}(\mathbf{W})$), and  $\mathbf{W}=\text{vec}^{-1}(\mathbf{w})$.

\subsection{Graph Smoothness}
In graph learning, a pivotal step is to elucidate how the topology of a network influences the behavior of signals defined on its nodes. A common assumption posits that such signals exhibit smooth variations over the graph \cite{kalofolias2016learn}, meaning that pairs of vertices linked by high-weight edges are expected to carry similar values. This concept of smoothness is formally captured by the quadratic form of the graph Laplacian. Given the Laplacian matrix $\mathbf{L}=\mathrm{Diag}(\mathbf{W1})-\mathbf{W}$, where $\mathbf{W}$ encodes the edge weights, the quantity
\begin{equation*}
    \mathbf{x}^T\mathbf{Lx}=\frac{1}{2}\sum_{i=1}^{m}\sum_{j=1}^mW_{ij}(x_i-x_j)^2,
\end{equation*}
serves as a scalar measure of how rapidly the signals $\mathbf{x}$ changes in accordance with the similarity structure in $\mathbf{W}$. We refer to this value as the ``local variation (LV)” of $\mathbf{x}$. For a collection of $n$ zero-mean graph signals assembled into the matrix $\mathbf{X}=[\mathbf{x}_1, \ldots, \mathbf{x}_n]\in\mathbb{R}^{m\times n}$, the aggregate LV is given by:
\begin{equation}
    \sum_{k=1}^n\mathbf{x}_k^T\mathbf{L}\mathbf{x}_k=\mathrm{tr}(\hat{\mathbf{C}}\mathbf{L}),
    \label{eq1}
\end{equation}
where $\hat{\mathbf{C}}:=\sum_{k=1}^n\mathbf{x}_k\mathbf{x}_k^T=\mathbf{XX}^T$. Additionally, dividing $\hat{\mathbf{C}}$ by $n$ yields the usual sample covariance matrix of the graph signals, under the simplifying assumption of zero mean.

\subsection{Influence of Partial Observations in the Topology Inference Model}

The current subsection is dedicated to formally delineating the topology-inference problem when observations are available only from a subset of nodes of the graph. We proffer a general formulation and underscore the impact of the concealed variables. 

Let $\mathbf{X}=[\mathbf{x}_1, \ldots, \mathbf{x}_n]\in \mathbb{R}^{m\times n}$ denote the compilation of $n$ signals superimposed on the unbeknownst graph $\mathcal{G}$ comprising $m$ nodes. We then posit that observations are available solely from a subset of nodes $\mathcal{O}\subset\mathcal{V}$ with cardinality $o<m$ and the remaining $h=m-o$ nodes are unobserved, denoted by the subset $\mathcal{H}=\mathcal{V}\backslash\mathcal{O}$. For the sake of simplicity and without compromising generality, assume $\mathcal{O}$ consists of the first $o$ nodes of the graph. Thus, the observed signal matrix $\mathbf{X}_{\mathcal{O}}\in\mathbb{R}^{o\times n}$ comprises the first $o$ rows of the matrix $\mathbf{X}$. We aim to discern the matrix $\hat{\mathbf{C}}\in\mathbb{R}^{m\times m}$ and the graph Laplacian $\mathbf{L}\in\mathbb{R}^{m\times m}$, which manifest the ensuing block configuration
\begin{align}
    \mathbf{X}=\begin{bmatrix}
        \mathbf{X}_{\mathcal{O}}\\
        \mathbf{X}_{\mathcal{H}}
    \end{bmatrix},
    \mathbf{L}=\begin{bmatrix}
        \mathbf{L}_{\mathcal{O}} & \mathbf{L}_{\mathcal{OH}}\\
        \mathbf{L}_{\mathcal{HO}} & \mathbf{L}_{\mathcal{H}}
    \end{bmatrix},
    \hat{\mathbf{C}}=\begin{bmatrix}
        \hat{\mathbf{C}}_{\mathcal{O}} & \hat{\mathbf{C}}_{\mathcal{OH}}\\
        \hat{\mathbf{C}}_{\mathcal{HO}} & \hat{\mathbf{C}}_{\mathcal{H}}
    \end{bmatrix}.
    \label{eq2}
\end{align}
The $o\times o$ matrix $\mathbf{L}_{\mathcal{O}}$ represents the Laplacian matrix of the observed nodes, whereas the residual blocks encapsulate the edges implicating concealed nodes. Analogously, $\frac{1}{n}\hat{\mathbf{C}}_{\mathcal{O}}=\frac{1}{n}\mathbf{X}_{\mathcal{O}}\mathbf{X}_{\mathcal{O}}^T$ denotes the sample covariance of the observed signals, while $1/n\hat{\mathbf{C}}_{\mathcal{OH}}$ and $1/n\hat{\mathbf{C}}_{\mathcal{HO}}$ correspond to the covariance involving hidden nodes. Given $\mathcal{G}$ is undirected, both $\mathbf{L}$ and $\hat{\mathbf{C}}$ are symmetric, thus, $\mathbf{L}_{\mathcal{HO}}=\mathbf{L}_{\mathcal{OH}}^T$ and $\hat{\mathbf{C}}_{\mathcal{HO}}=\hat{\mathbf{C}}_{\mathcal{OH}}^T$.

 The graph learning/network topology inference problem under partial observations is formally introduced subsequently:

\textbf{\textit{Problem 1:}} Given a graph $\mathcal{G}=(\mathcal{V},\mathcal{E})$ with $m$ nodes and an unknown Laplacian matrix $\mathbf{L}\in\mathbb{R}^{m\times m}$, and the observations $\mathbf{X}_{\mathcal{O}}\in\mathbb{R}^{o\times n}$ corresponding to the values of $n$ graph signals at the nodes in $\mathcal{O}$ with $|\mathcal{O}|=o$, discern the underlying graph structure encoded $\mathbf{L}_{\mathcal{O}}\in\mathbb{R}^{o\times o}$ under the assumptions that:

(\textbf{A1}) The number of hidden nodes is much smaller than the number of observed nodes, i.e., $h\ll o$;

(\textbf{A2}) Graph signals $\mathbf{X}\in\mathbb{R}^{m\times n}$ varies smoothly across the graph.

\begin{remark}
    Despite possessing observations from $o$ nodes, the presence of $h=m-o$ hidden nodes complicates the inference of $\mathbf{L}_{\mathcal{O}}$, making the problem challenging, (\textbf{A1}) simplifies the problem by limiting the number of concealed variables. Assumption (\textbf{A2}) ensures graph signals $\mathbf{X}$ varies smoothly across the graph, with Equation \eqref{eq1} quantifying the smoothness of graph signals $\mathbf{X}$ on the graph $\mathcal{G}$, which provides a relationship between the graph signals and the graph Laplacian matrix $\mathbf{L}$.
\end{remark}

\begin{remark}
    In real-world scenarios, the Assumption (\textbf{A1}) that the number of observed nodes substantially exceeds that of unobserved nodes is indeed justifiable in certain contexts. For example, in sensor networks deployed for environmental monitoring or structural health evaluation, sensors are often densely and strategically placed. This high sensor density means that most of the network is instrumented, and only a small subset of nodes may be missing or faulty \cite{chen2022adaptive}. Similarly, in engineered systems such as power grids or communication networks, extensive monitoring infrastructures typically ensure comprehensive coverage of nodes, with unobserved nodes representing only a minimal fraction \cite{ramakrishna2021grid}. These conditions facilitate more accurate inference of the underlying graph topology, thereby validating the assumption adopted in many theoretical formulations of network topology inference\cite{buciulea2022learning}, \cite{10507157}. However, it is important to recognize that this assumption may not be universally applicable. In domains like social networks or certain biological systems, the proportion of unobserved nodes can be significant due to privacy constraints, data collection limitations, or intrinsic network complexity \cite{sia2022inferring}, \cite{ma2023single}. Thus, while the assumption holds in many engineered and controlled environments, caution is warranted when extending it to more complex or less accessible networks
\end{remark}

\subsection{Problem Formulation of Topology Inference from Smooth Signals under Partial Observations}
 As previously discussed, to measure the smoothness of graph signals, we utilize the graph Laplacian and evaluate their LV as $\mathrm{tr}(\mathbf{X}^T\mathbf{L}\mathbf{X})$. Here, we focus on $\hat{\mathbf{C}}=\mathbf{XX}^T$. Due to the presence of hidden nodes, the entire covariance matrix is unobservable. To handle this, capitalizing on the block structure of $\hat{\mathbf{C}}$ and $\mathbf{L}$ introduced in Equation \eqref{eq2}, we can reformulate the LV of our dataset as
\begin{equation}
    \mathrm{tr}(\hat{\mathbf{C}}\mathbf{L})=\mathrm{tr}(\hat{\mathbf{C}}_{\mathcal{O}}\mathbf{L}_{\mathcal{O}})+2\mathrm{tr}(\hat{\mathbf{C}}_{\mathcal{OH}}\mathbf{L}_{\mathcal{OH}}^T)+\mathrm{tr}(\hat{\mathbf{C}}_{\mathcal{H}}\mathbf{L}_{\mathcal{H}}),
    \label{eq3}
\end{equation}
where only $\hat{\mathbf{C}}_{\mathcal{O}}=\mathbf{X}_{\mathcal{O}}\mathbf{X}_{\mathcal{O}}^T$ presumed to be known and the impact of hidden variables in the LV has been elucidated.

While the block-wise smoothness expression in \eqref{eq3} could be directly employed to approach the network topology inference as an optimization problem, the unknown submatrices  necessitate estimation, making the problem non-convex if terms $\hat{\mathbf{C}}_{\mathcal{OH}}\mathbf{L}_{\mathcal{OH}}^T$ and $\hat{\mathbf{C}}_{\mathcal{H}}\mathbf{L}_{\mathcal{H}}$ are included. To circumvent this predicament, \cite{buciulea2022learning} introduce matrix $\mathbf{K}:=\hat{\mathbf{C}}_{\mathcal{OH}}\mathbf{L}_{\mathcal{OH}}^T\in\mathbb{R}^{o\times o}$, leveraging its low-rank property due to $\mathrm{rank}(\mathbf{K})\leq h\ll o$. Similarly, matrix $\mathbf{R}:=\hat{\mathbf{C}}_{\mathcal{H}}\mathbf{L}_{\mathcal{H}}\in\mathbb{R}^{h\times h}$ is a positive semi-definite matrices\footnote{Although in practical applications we might not know the exact value of $h$, the choice of $h$ does not affect the formulation and solution of the subsequent optimization problem.}, ensuring $\mathrm{tr}(\mathbf{R})\geq 0$.

 Thus, similar to \cite{buciulea2022learning}, the network topology inference from smooth signals under partial observations is formulated as 
 \begin{equation}
     \begin{aligned}
         & \min_{\mathbf{L}_{\mathcal{O}}, \mathbf{K}, \mathbf{R}} \mathrm{tr}(\hat{\mathbf{C}}_{\mathcal{O}}\mathbf{L}_{\mathcal{O}})+2\mathrm{tr}(\mathbf{K})+\mathrm{tr}(\mathbf{R})-\alpha\mathbf{1}^T\log(\mathrm{diag}(\mathbf{L}_{\mathcal{O}}))+\beta\Vert\mathbf{L}_{\mathcal{O}}\Vert_{F,off}^2+\gamma\Vert\mathbf{K}\Vert_{\ast}\\
         &\quad\begin{array}{r@{\quad}r@{}l@{\quad}l}
            \mathrm{s.t.} &\mathrm{tr}(\hat{\mathbf{C}}_{\mathcal{O}}\mathbf{L}_{\mathcal{O}})+2\mathrm{tr}(\mathbf{K})+\mathrm{tr}(\mathbf{R}) &\geq 0,\\
     &\mathrm{tr}(\mathbf{R})\geq 0, \mathbf{L}_{\mathcal{O}} &\in\bar{\mathcal{L}},
         \end{array}
     \end{aligned}
     \label{eq4}
 \end{equation}
 where $\Vert\cdot\Vert_{F,off}$ denotes the Forbenius norm excluding the elements of the diagonal. This term, in conjunction with $\log(\mathrm{diag}(\mathbf{L}_{\mathcal{O}}))$, controls the sparsity of $\mathbf{L}_{\mathcal{O}}$ and precludes the trivial solution of $\mathbf{L}_{\mathcal{O}}=\mathbf{0}$. The nuclear norm $\Vert\cdot\Vert_{\ast}$ encourages low-rank solutions for $\mathbf{K}$, and its incorporation ensure the convexity of \eqref{eq4}, allowing for a globally optimal solution. The parameters $\alpha, \beta, \gamma\geq 0$ balance the regularizers, with the first constraint ensuring non-negativity of LV and the second constraint encapsulating the positive semi-definiteness of $\mathbf{R}$. The final consideration is the form of $\bar{\mathcal{L}}$. It equates to the set of combinatorial Laplacians $\mathcal{L}$, but substituting the condition $\mathbf{L}\mathbf{1}=\mathbf{0}$ with $\mathbf{L}\mathbf{1}\geq\mathbf{0}$, i.e., $\bar{\mathcal{L}}:=\{\mathbf{L}\in\mathbb{R}^{m\times m}\mid L_{ij}\leq 0, \text{for} \quad i\neq j; \mathbf{L}=\mathbf{L}^T; \mathbf{L}\mathbf{1}\geq\mathbf{0}; \mathbf{L}\succeq 0\}$, as $\mathbf{L}_{\mathcal{O}}$ is not a combinatorial Laplacian due to connections between observed and hidden nodes. This relaxation, while expanding the set of feasible solutions, complicates inference. To mitigate this, we shift to estimating $\tilde{\mathbf{L}}_{\mathcal{O}}:=\mathrm{Diag}(\mathbf{W}_{\mathcal{O}}\mathbf{1})-\mathbf{W}_{\mathcal{O}}$, the Laplacian associated with the observed adjacency matrix $\mathbf{W}_{\mathcal{O}}\in\mathbb{R}^{o\times o}$ in which $\mathbf{W}_{\mathcal{O}}$ is the upper-left block of the partitioned matrix $\mathbf{W}$, i.e., similar to \eqref{eq2}
\begin{align*}
    \mathbf{W} = \begin{bmatrix}
        \mathbf{W}_\mathcal{O} & \mathbf{W}_\mathcal{OH}\\
        \mathbf{W}_\mathcal{HO} & \mathbf{W}_\mathcal{H}
    \end{bmatrix}.
\end{align*}
Unlike $\mathbf{L}_{\mathcal{O}}$, $\tilde{\mathbf{L}}_{\mathcal{O}}$ is a valid combinatorial Laplacian, allowing reinstatement of the original Laplacian constraints. 

Upon defining the $o\times o$ diagonal matrices $\mathbf{D}_{\mathcal{O}}:=\mathrm{Diag}(\mathbf{W}_{\mathcal{O}}\mathbf{1})$ and $\mathbf{D}_{\mathcal{OH}}:=\mathrm{Diag}(\mathbf{W}_{\mathcal{OH}}\mathbf{1})$, which enumerate the number of observed and concealed neighbors for the nodes in $\mathcal{O}$, the matrix $\mathbf{L}_{\mathcal{O}}$ is expressed as $\mathbf{L}_{\mathcal{O}}=\mathbf{D}_{\mathcal{O}}+\mathbf{D}_{\mathcal{OH}}-\mathbf{W}_{\mathcal{O}}=\tilde{\mathbf{L}}_{\mathcal{O}}+\mathbf{D}_{\mathcal{OH}}$. With this equivalence, the smoothness penalty in \eqref{eq3} is rewritten as
\begin{align*}
    \mathrm{tr}(\hat{\mathbf{C}}\mathbf{L}) &=\mathrm{tr}(\hat{\mathbf{C}}_{\mathcal{O}}\tilde{\mathbf{L}}_{\mathcal{O}})+\mathrm{tr}(\hat{\mathbf{C}}_{\mathcal{O}}\mathbf{D}_{\mathcal{OH}})+2\mathrm{tr}(\mathbf{K})+\mathrm{tr}(\mathbf{R})\\
    &= \mathrm{tr}(\hat{\mathbf{C}}_{\mathcal{O}}\tilde{\mathbf{L}}_{\mathcal{O}})+2\mathrm{tr}(\tilde{\mathbf{K}})+\mathrm{tr}(\mathbf{R}),
\end{align*}
 where $\tilde{\mathbf{K}}:=\hat{\mathbf{C}}_{\mathcal{O}}\mathbf{D}_{\mathcal{OH}}/2+\mathbf{K}$. Owing to the entries of $\mathbf{D}_{\mathcal{OH}}$ depend on the presence of edges between the observed and the hidden nodes, if the graph is sparse, the matrix $\mathbf{D}_{\mathcal{OH}}$ will exhibit low-rank characteristics. Furthermore, since the sparsity pattern of the diagonal $\mathbf{D}_{\mathcal{OH}}$ is contingent upon the matrix $\mathbf{W}_{\mathcal{OH}}=-\mathbf{L}_{\mathcal{OH}}$, it follows that the column sparsity pattern of $\hat{\mathbf{C}}_{\mathcal{O}}\mathbf{D}_{\mathcal{OH}}$ aligns with that of $\mathbf{K}$, thereby implying that $\tilde{\mathbf{K}}$ also possesses low rank properties.

The optimization in \eqref{eq4} is then recast as 
\begin{equation}
     \begin{aligned}
         & \min_{\tilde{\mathbf{L}}_{\mathcal{O}}, \tilde{\mathbf{K}}, r} \mathrm{tr}(\hat{\mathbf{C}}_{\mathcal{O}}\tilde{\mathbf{L}}_{\mathcal{O}})+2\mathrm{tr}(\tilde{\mathbf{K}})+r-\alpha\mathbf{1}^T\log(\mathrm{diag}(\tilde{\mathbf{L}}_{\mathcal{O}}))+\beta\Vert\tilde{\mathbf{L}}_{\mathcal{O}}\Vert_{F,off}^2+\gamma_{\ast}\Vert\tilde{\mathbf{K}}\Vert_{\ast}+\gamma_{2,1}\Vert\tilde{\mathbf{K}}\Vert_{2,1}\\
         &\quad\begin{array}{r@{\quad}r@{}l@{\quad}l}
            \mathrm{s.t.} &\mathrm{tr}(\hat{\mathbf{C}}_{\mathcal{O}}\tilde{\mathbf{L}}_{\mathcal{O}})+2\mathrm{tr}(\tilde{\mathbf{K}})+r &\geq 0,\\
     &r\geq 0, \tilde{\mathbf{L}}_{\mathcal{O}} &\in\mathcal{L},
         \end{array}
     \end{aligned}
     \label{eq5}
 \end{equation}
 with $\bar{\mathcal{L}}$ in \eqref{eq4} has been substituted with $\mathcal{L}$ in \eqref{eq5}, which represents the set of all valid combinatorial Laplacian matrices. Matrix $\mathbf{R}$ is replaced by the non-negative variable $r$ to simplify computation. Although we replaced $\mathbf{K}$ with $\tilde{\mathbf{K}}$, it retains the same regularization properties. This formulation ensures a solution with the desired column sparsity and low-rank characteristics, with the flexibility to use either the nuclear norm or group LASSO penalty based on $\gamma_{2,1}$ and $\gamma_\ast$ \cite{buciulea2022learning}.

\section{Proposed Optimization Framework}\label{sect3}
\subsection{Problem Reformulation}
In this context, let $\tilde{\mathbf{x}}_i:=[(\mathbf{x}_1)_i, \ldots, (\mathbf{x}_n)_i]^T$ denote the data vector associated with the $i$-th node and define $Z_{ij}:=\Vert\tilde{\mathbf{x}}_i-\tilde{\mathbf{x}}_j\Vert_2^2$ as the squared pair-wise distance between the node vectors $\tilde{\mathbf{x}}_i$, $\tilde{\mathbf{x}}_j$. Subsequently, we derive the following expression, commonly recognized as the Dirichlet energy:
\begin{align*}
    \sum_{k=1}^n\mathbf{x}_k^T\mathbf{Lx}_k&=\mathrm{tr}(\hat{\mathbf{C}}\mathbf{L})\\
    &=\frac{1}{2}\sum_{i=1}^m\sum_{j=1}^mW_{ij}\Vert\tilde{\mathbf{x}}_i-\tilde{\mathbf{x}}_j\Vert_2^2\\
    &=\frac{1}{2}\Vert\mathbf{W}\odot\mathbf{Z}\Vert_{1, 1}.
\end{align*}

Thus, equivalently, optimization problem \eqref{eq5} can be reformulated as\footnote{In Problem $1$, we need to find the graph Laplacian matrix $\mathbf{L}_{\mathcal{O}}$. Although in \eqref{eq6}, we are finding the adjacency matrix $\mathbf{W}_\mathcal{O}$, since $\mathbf{L}_\mathcal{O}=\text{Diag}(\mathbf{W}_\mathcal{O}\mathbf{1})-\mathbf{W}_\mathcal{O}$, if we can determine $\mathbf{W}_\mathcal{O}$, then $\mathbf{L}_\mathcal{O}$ can also be obtained.}
\begin{equation}
    \begin{aligned}
        & \min_{\mathbf{W}_\mathcal{O}\in\mathbb{R}^{o\times o}, \tilde{\mathbf{K}}, r} \frac{1}{2}\Vert\mathbf{W}_{\mathcal{O}}\odot\mathbf{Z}_{\mathcal{O}}\Vert_{1,1}+2\mathrm{tr}(\tilde{\mathbf{K}})+r+\beta\Vert\mathbf{W}_{\mathcal{O}}\Vert_{F}^2-\alpha\mathbf{1}^T\log(\mathbf{W}_{\mathcal{O}}\mathbf{1})+\gamma_{\ast}\Vert\tilde{\mathbf{K}}\Vert_{\ast}+\gamma_{2,1}\Vert\tilde{\mathbf{K}}\Vert_{2,1}\\
        &\qquad\begin{array}{r@{\quad}r@{}l@{\quad}l}
            \mathrm{s.t.} &\mathbf{W}_\mathcal{O}\geq 0, \mathbf{W}_{\mathcal{O}}=\mathbf{W}_{\mathcal{O}}^T, \mathrm{diag}(\mathbf{W}_{\mathcal{O}})&=\mathbf{0},\\
            &\frac{1}{2}\Vert\mathbf{W}_{\mathcal{O}}\odot\mathbf{Z}_{\mathcal{O}}\Vert_{1,1}+2\mathrm{tr}(\tilde{\mathbf{K}})+r &\geq 0,\\
             &r &\geq 0,\\
        \end{array}
    \end{aligned}
    \label{eq6}
\end{equation}
where $\mathbf{Z}_{\mathcal{O}}$ is the upper-left block of the partitioned matrix $\mathbf{Z}$ like \eqref{eq2}.

Indeed, two distinct configurations of the objective function emerge based on the selection of regularization constants. By setting $\gamma_\ast = 0$, we encourage a solution that exhibits the desired column-sparsity property of the matrix $\tilde{\mathbf{K}}$. Conversely, setting $\gamma_{2,1} = 0$ leads to the promotion of a solution characterized by a reduced rank of $\tilde{\mathbf{K}}$ achieved through the imposition of nuclear norm regularization. This strategic manipulation of regularization parameters allows for the fine-tuning of the structural properties of $\tilde{\mathbf{K}}$ thereby enabling targeted optimization according to specific graph learning objectives.

Initially, we consider the scenario where $\gamma_\ast=0$. Let $\mathbf{w}$ (resp. $\mathbf{z}$) be the vector constructed by concatenating the entries above the main diagonal of $\mathbf{W}_{\mathcal{O}}$ (resp. $\mathbf{Z}_{\mathcal{O}}$) into a single column. Consequently, we have $\mathbf{w}\in\mathbb{R}^{p}$ and $\mathbf{z}\in\mathbb{R}^{p}$ with $p:=o(o-1)/2$. Similarly, we define $\mathbf{k}:=\text{vec}(\tilde{\mathbf{K}})\in\mathbb{R}^{o^2}$. Thus, we can reformulate equation \eqref{eq6} as (with $\gamma_\ast=0$)
\begin{equation}
    \begin{aligned}
        &\min_{\mathbf{w}\in\mathbb{R}^p, \mathbf{k}\in\mathbb{R}^{o^2}, r\in\mathbb{R}} \frac{1}{2}\mathbf{z}^T\mathbf{w}+2\mathbf{b}^T\mathbf{k}+r-\alpha\mathbf{1}^T\log(\mathbf{Bw})+\beta\Vert\mathbf{w}\Vert_2^2+\gamma_{2,1}h(\mathbf{k})\\
        &\qquad\begin{array}{r@{\quad}r@{}l@{\quad}l}
            \mathrm{s.t.} &\mathbf{w}\geq 0, r &\geq 0\\
            & \frac{1}{2}\mathbf{z}^T\mathbf{w}+2\mathbf{b}^T\mathbf{k}+r &\geq 0\\
        \end{array}
    \end{aligned}
    \label{eq7}
\end{equation}
where $h(\mathbf{k}):=\Vert\tilde{\mathbf{K}}\Vert_{2,1}$, and $\mathbf{B}\in\{0, 1\}^{m\times p}$ and $\mathbf{b}\in\{0, 1\}^{o^2}$ are defined such that:
\begin{align}
    \mathbf{Bw}&=\mathbf{W1},
    \label{eq8}\\
    \mathbf{b}^T\mathbf{k}&=\mathrm{tr}(\tilde{\mathbf{K}}).
    \label{eq9}
\end{align}

Through the introduction of an auxiliary variable $v\in\mathbb{R}_{+}$, and by setting $\frac{1}{2}\mathbf{z}^T\mathbf{w}+2\mathbf{b}^T\mathbf{k}+r-v=0$. Further refinement is obtained by introducing new variables $\mathbf{v}:=[v_1;v_2]$ with $v_1:=r\in\mathbb{R}$, $v_2\:=v\in\mathbb{R}$, new variables $\mathbf{u}$ such that $\mathbf{Bw}=\mathbf{u}$ and denoting $\mathbf{a}:=[1\quad -1]^T, \mathbf{d}:=[1\quad 0]^T\in\mathbb{R}^2$. This leads to a subsequent reformulation of problem \eqref{eq7}:
\begin{equation}
    \begin{aligned}
        &\min_{\mathbf{w},\mathbf{u}, \mathbf{k}, \mathbf{v}} f_1(\mathbf{w})+f_2(\mathbf{u})+f_3(\mathbf{k})+f_4(\mathbf{v})\\
        &\begin{array}{r@{\quad}r@{}l@{\quad}l}
            \mathrm{s.t.}\qquad&\frac{1}{2}\mathbf{z}^T\mathbf{w}+2\mathbf{b}^T\mathbf{k}+\mathbf{a}^T\mathbf{v}&=0\\
            &\mathbf{Bw}&=\mathbf{u}.
        \end{array}
    \end{aligned}
    \label{eq10}
\end{equation}
where the functions are redefined as:
\begin{align*}
    f_1(\mathbf{w}) &:= \frac{1}{2}\mathbf{z}^T\mathbf{w}+\beta\Vert\mathbf{w}\Vert_2^2+\mathbb{I}_{\mathbb{R}_+^p}(\mathbf{w}),
    \quad f_2(\mathbf{u}) :=-\alpha\mathbf{1}^T\log(\mathbf{u}),\\
    f_3(\mathbf{k}) &:= 2\mathbf{b}^T\mathbf{k}+\gamma_{2,1}h(\mathbf{k}),
    \quad f_4(\mathbf{v}) :=\mathbf{d}^T\mathbf{v}+\mathbb{I}_{\mathbb{R}_+}(\mathbf{v}),
\end{align*}
where $h(\mathbf{k})=\Vert\tilde{\mathbf{K}}\Vert_{2,1}$. 

Denoting $\mathbf{x}:=[\mathbf{w};\mathbf{u};\mathbf{k};\mathbf{v}]$, Problem \eqref{eq10} can be reformulated as
\begin{equation}
    \begin{aligned}
        & \min_{\mathbf{x}} f(\mathbf{x})=f_1(\mathbf{w})+f_2(\mathbf{u})+f_3(\mathbf{k})+f_4(\mathbf{v})\\
        &\begin{array}{r@{\quad}r@{}l@{\quad}l}
            \mathrm{s.t.}\qquad \mathbf{Ax}=\mathbf{0}.
        \end{array}
    \end{aligned}
    \label{eq11}
\end{equation}
where,
\begin{align}
    f_1(\mathbf{w}) &:= \frac{1}{2}\mathbf{z}^T\mathbf{w}+\beta\Vert\mathbf{w}\Vert_2^2+\mathbb{I}_{\mathbb{R}_+^p}(\mathbf{w}), \quad
    f_2(\mathbf{u}) :=-\alpha\mathbf{1}^T\log(\mathbf{u}),\nonumber\\
    f_3(\mathbf{k}) &:= 2\mathbf{b}^T\mathbf{k}+\gamma_{2,1}h(\mathbf{k}),\quad f_4(\mathbf{v}) :=\mathbf{d}^T\mathbf{v}+\mathbb{I}_{\mathbb{R}_+}(\mathbf{v}),\nonumber\\
    \mathbf{A}&:=\begin{bmatrix}
        \frac{1}{2}\mathbf{z}^T & \mathbf{0}& 2\mathbf{b}^T & \mathbf{a}^T \\
        \mathbf{B} &-\mathbf{I} & \mathbf{0} & \mathbf{0} 
    \end{bmatrix}.
    \nonumber
\end{align}
Furthermore, we denote $\mathbf{M}_1=[1/2\mathbf{z}^T;\mathbf{B}]\in\mathbb{R}^{(1+o)\times p}$, $\mathbf{M}_2=[\mathbf{0};-\mathbf{I}]\in\mathbb{R}^{(1+o)\times o}$, $\mathbf{M}_3=[2\mathbf{b}^T;\mathbf{0}]\in\mathbb{R}^{(1+o)\times o^2}$ and $\mathbf{M}_4=[\mathbf{a}^T;\mathbf{0}]\in\mathbb{R}^{(1+o)\times 2}$, so $\mathbf{A}=[\mathbf{M}_1^T;\mathbf{M}_2^T;\mathbf{M}_3^T;\mathbf{M}_4^T]^T$.

\begin{remark}
    Although in practical applications we might not know the exact value of $h$, the choice of $h$ does not affect the formulation and solution of our optimization problem. The reason is as follows: Originally, the matrices $\hat{\mathbf{C}}_{\mathcal{OH}}$,  $\mathbf{L}_{\mathcal{OH}}$ and $\mathbf{R}$ depend on $h$. However, by introducing a new matrix $\mathbf{K}:=\hat{\mathbf{C}}_{\mathcal{OH}}\mathbf{L}_{\mathcal{OH}}^T\in\mathbb{R}^{o\times o}$, we make the matrix $\mathbf{K}$ independent of $h$. Furthermore, although $\mathbf{R}$ is a $h$-dimensional square matrix, our optimization objective only involves the trace of $\mathbf{R}$. To reduce computational and storage burdens, we substitute $\text{tr}(\mathbf{R})$ with a one-dimensional variable $r$. This substitution ensures that the choice of 
$h$ does not affect the solution to our optimization problem \eqref{eq11}.
\end{remark}

\subsection{Algorithmic Development}
To begin, let $\lambda=[\lambda_1;\lambda_2]\in\mathbb{R}^{o+1}$ be the dual variable associated with the linear constraints $\mathbf{Ax}=\mathbf{0}$ in Problem \eqref{eq11}, thus we can obtain the augmented Lagrangian function of the form
\begin{align}
    \mathcal{L}(\mathbf{x};\lambda)=f(\mathbf{x})-\lambda^T(\mathbf{Ax})+\frac{\rho}{2}\Vert\mathbf{Ax}\Vert_2^2,
    \label{eq12}
\end{align}
where $\rho\geq\mathbf{0}$ is a constant. The augmented dual function is given by 
\begin{align}
    d(\lambda)=\min_{\mathbf{x}}f(\mathbf{x})-\lambda^T\mathbf{Ax}+\frac{\rho}{2}\Vert\mathbf{Ax}\Vert_2^2
    \label{eq13}
\end{align}

In the $i$-th iteration (where $i\geq 0$), our GLOPSS-CS proceeds with the updates
\begin{align}
    \mathbf{w}^{i+1}&=\mathrm{arg}\min_{\mathbf{w}\in\mathbb{R}^p}\mathcal{L}(\mathbf{w}, \mathbf{k}^i, \mathbf{v}^i, \mathbf{u}^i; \lambda^i), \nonumber\\
    \mathbf{u}^{i+1}&=\mathrm{arg}\min_{\mathbf{u}\in\mathbb{R}^o}\mathcal{L}(\mathbf{w}^{i+1}, \mathbf{u},\mathbf{k}^i, \mathbf{v}^i; \lambda^i), \nonumber\\
    \mathbf{k}^{i+1}&=\mathrm{arg}\min_{\mathbf{k}\in\mathbb{R}^{o^2}}\mathcal{L}(\mathbf{w}^{i+1}, \mathbf{u}^{i+1}, \mathbf{k}, \mathbf{v}^i; \lambda^i), \nonumber\\
    \mathbf{v}^{i+1}&=\mathrm{arg}\min_{\mathbf{v}\in\mathbb{R}^2}\mathcal{L}(\mathbf{w}^{i+1}, \mathbf{u}^{i+1}, \mathbf{k}^{i+1}, \mathbf{v}; \lambda^i), \nonumber\\
    \lambda^{i+1}&=\lambda^i-\rho\mathbf{Ax}^{i+1}.
    \label{eq14}
\end{align}

\subsection{\texorpdfstring{$\mathbf{w}$}{w}-Subproblem}
We first provide the following proposition as preparation, with its proof deferred to the \ref{app:prop1}.
\begin{prop}\label{prop1}
    If $f_{1}(\mathbf{w})=\frac{1}{2}\mathbf{z}^T\mathbf{w}+\beta\Vert\mathbf{w}\Vert_2^2+\mathbb{I}_{\mathbb{R}_+^p}(\mathbf{w})$, then for $\tau_1>0$, the closed-form proximal mapping of $f_{1}$ is given by
    \begin{equation*}
        \mathrm{prox}_{\tau_1f_{1}}(\mathbf{w})=\max\left\{\frac{\mathbf{w}-\frac{1}{2}\tau_1\mathbf{z}}{2\tau_1\beta+1}, \mathbf{0}\right\}.
    \end{equation*}
\end{prop}

Fixing $\mathbf{k}^i$, $\mathbf{v}^i$, $\mathbf{u}^i$ and $\lambda^i$ in the $i$-th iteration, the subproblem for $\mathbf{w}$ is
\begin{align}
    &\min_{\mathbf{w}} f_{1}(\mathbf{w})-(\lambda^i)^T\mathbf{A}[\mathbf{w};\mathbf{k}^i;\mathbf{v}^i;\mathbf{u}^i]+\frac{\rho}{2}\Vert\mathbf{A}[\mathbf{w};\mathbf{k}^i;\mathbf{v}^i;\mathbf{u}^i]\Vert_2^2 \nonumber\\
    &=\min_{\mathbf{w}} f_1(\mathbf{w})+\frac{\rho}{2}\Vert\mathbf{A}[\mathbf{w};\mathbf{k}^i;\mathbf{v}^i;\mathbf{u}^i]-\frac{\lambda^i}{\rho}\Vert_2^2
    \label{eq15}
\end{align}
In view of Proposition \ref{prop1}, we linearize the quadratic term to perform one step of the proximal gradient iteration to update $\mathbf{w}$
\begin{align}
    \mathbf{w}^{i+1} &=\mathrm{prox}_{\tau_1f_{1}}(\mathbf{w}^i-\tau_1\mathbf{e}_1^T\mathbf{A}^T(\mathbf{Ax}^i-\frac{\lambda^i}{\rho})) \nonumber\\
    &=\max\{\tilde{\mathbf{w}}^{i+1}, \mathbf{0}\},
    \label{eq16}
\end{align}
where $\mathbf{e}_1=[1;0;0;0]^T\in\mathbb{R}^4$ and
\begin{align}
    \tilde{\mathbf{w}}^{i+1}=\frac{\mathbf{w}^i-\tau_1\mathbf{B}^T(\mathbf{Bw}^i+\omega_2^i)+\omega_3^i-\frac{1}{4}\tau_1\rho\mathbf{zz}^T}
    {2\tau_1\beta+1}.
    \nonumber
\end{align}
Here, $\omega_1^i=2\mathbf{b}^T\mathbf{k}^i+\mathbf{a}^T\mathbf{v}^i-\frac{\lambda_1^i}{\rho}$ is a constant, $\omega_2^i=-\mathbf{u}^i-\frac{\lambda_2^i}{\rho}\in\mathbb{R}^o$, and $\omega_3^i=-\frac{1}{2}\tau_1\mathbf{z}-\frac{1}{2}\tau_1\rho\omega_1^i\mathbf{z}$.

\subsection{\texorpdfstring{$\mathbf{u}$}{u}-Subproblem}
We first provide the following lemma, adapted from \cite{parikh2014proximal}, as preparation.
\begin{lem}\label{lemma1}
    If $f_2(\mathbf{u})=-\alpha\mathbf{1}^T\log(\mathbf{u})$, then for $\tau_2>0$, the closed-form proximal mapping of $f_{2}$ is given by
    \begin{equation*}
        \mathrm{prox}_{\tau_2f_{2}}(\mathbf{u})=\frac{\mathbf{u}+\sqrt{\mathbf{u}^2+4\alpha\tau_2\mathbf{1}}}{2},
    \end{equation*}
    where the square and the square root are both taken element-wise.
\end{lem}

Fixing $\lambda^i$ and the newly updated $\mathbf{w}^{i+1}$, the subproblem for $\mathbf{u}$ is
\begin{align}
    \min_{\mathbf{u}} f_{2}(\mathbf{u})+\frac{\rho}{2}\Vert\mathbf{A}[\mathbf{w}^{i+1};\mathbf{k}^i;\mathbf{v}^i;\mathbf{u}]-\frac{\lambda^i}{\rho}\Vert_2^2.
    \label{eq17}
\end{align}
Now, in view of Lemma \ref{lemma1}, we perform similar proximal gradient iteration to update $\mathbf{u}$
\begin{align}
    \mathbf{u}^{i+1} &=\mathrm{prox}_{\tau_2f_{2}}(\mathbf{u}^i-\tau_2\mathbf{e}_2^T\mathbf{A}^T(\mathbf{Ax}^i-\frac{\lambda^i}{\rho})) \nonumber\\
    &=\frac{\tilde{\mathbf{u}}^{i+1}+\sqrt{(\tilde{\mathbf{u}}^{i+1})^2+4\alpha\tau_2\mathbf{1}}}{2},
    \label{eq18}
\end{align}
where $\mathbf{e}_2=[0;1;0;0]^T\in\mathbb{R}^4$ and $\tilde{\mathbf{u}}^{i+1}=(1-\tau_2\rho)\mathbf{u}^i+\tau_2\rho\mathbf{Bw}^{i+1}-\tau_2\lambda_2^i/\rho$. 

\subsection{\texorpdfstring{$\mathbf{k}$}{k}-Subproblem}
Similarly, before we give the update of $\mathbf{k}$, we provide a lemma adapted from \cite{beck2017first} as preparation.
\begin{lem}\label{lemma2}
    Let $f:\mathbb{E}\rightarrow\mathbb{R}$ be a function given by $f(\mathbf{x}):=g(\Vert\mathbf{x}\Vert_2)$, where $g:\mathbb{R}\rightarrow(-\infty, +\infty]$ is a proper closed and convex function satisfying $\mathrm{dom}(g)\subseteq[0, +\infty)$. Then
    \begin{equation*}
        \mathrm{prox}_{\lambda f}(\mathbf{x})=\left\{
        \begin{aligned}
            &\mathrm{prox}_{\lambda g}(\Vert\mathbf{x}\Vert_2)\frac{\mathbf{x}}{\Vert\mathbf{x}\Vert_2},\qquad\qquad &\mathbf{x}\neq \mathbf{0},\\
            &\{\mathbf{u}\in\mathbb{E}:\Vert\mathbf{u}\Vert_2=\mathrm{prox}_{\lambda g}(\mathbf{0})\}, &\mathbf{x=0}.
        \end{aligned}
        \right.
    \end{equation*}
\end{lem}
Combining Lemma \ref{lemma2}, we can immediately have the following proposition, showing simple closed forms of the proximal mapping of $f_3$, with its proof deferred to the \ref{app:prop2}:
\begin{prop}\label{prop2}
     If $f_3(\mathbf{k})=2\mathbf{b}^T\mathbf{k}+\gamma_{2,1}h(\mathbf{k})$, then for $\tau_3>0$, the closed-form proximal mapping of $f_{3}$ is given by
     \begin{equation*}
         \mathrm{prox}_{\tau_3f_3}(\mathbf{k})=\left[1-\frac{\tau_3\gamma_{2,1}}{\Vert\mathbf{k}-2\tau_3\mathbf{b}\Vert_2}\right]_+(\mathbf{k}-2\tau_3\mathbf{b}),
     \end{equation*}
     if $\Vert\mathbf{k}\Vert_2\neq 0$, else $0$.
\end{prop}

Fixing $\mathbf{w}^{i+1}$, $\mathbf{u}^{i+1}$, $\mathbf{v}^i$ and $\lambda^i$, the subproblem for $\mathbf{k}$ is
\begin{align}
    \mathbf{k}^{i+1}&=\mathrm{arg}\min_{\mathbf{k}\in\mathbb{R}^{o^2}}\mathcal{L}(\mathbf{w}^{i+1}, \mathbf{u}^{i+1}, \mathbf{k}, \mathbf{v}^i; \lambda^i) \nonumber\\
    &=\text{arg}\min_{\mathbf{k}}f_3(\mathbf{k})+\frac{\rho}{2}\Vert\mathbf{A}[\mathbf{w}^{i+1};\mathbf{k};\mathbf{v}^i;\mathbf{u}^{i+1}]-\frac{\lambda^i}{\rho}\Vert_2^2.
    \label{eq19}
\end{align}
In view of Proposition \ref{prop2}, we linearize the quadratic term to perform one step of the proximal gradient iteration to update $\mathbf{k}$
\begin{align}
    \mathbf{k}^{i+1} & = \mathrm{prox}_{\tau_3f_{3}}(\mathbf{k}^i-\tau_3\mathbf{e}_3^T\mathbf{A}^T(\mathbf{Ax}^i-\frac{\lambda^i}{\rho}) \nonumber\\
    &= \left[1-\frac{\tau_3\gamma_{2,1}}{\Vert\tilde{\mathbf{k}}^{i+1}\Vert_2}\right]_+\tilde{\mathbf{k}}^{i+1},
    \label{eq20}
\end{align}
where $\mathbf{e}_3=[0;0;1;0]^T\in\mathbb{R}^4$ and
\begin{align}
    \tilde{\mathbf{k}}^{i+1}=&\mathbf{k}^i-\tau_32(1+\frac{1}{2}\rho\mathbf{z}^T\mathbf{w}^{i+1} +2\rho\mathbf{b}^T\mathbf{k}^i+\rho\mathbf{a}^T\mathbf{v}^{i}-\lambda_1^i)\mathbf{b}.
    \nonumber
\end{align}
Here, $[\mathbf{a}]_+=\max\{\mathbf{a}, \mathbf{0}\}$ denotes the element-wise maximum of $\mathbf{a}$ and $\mathbf{0}$.

\subsection{\texorpdfstring{$\mathbf{v}$}{v}-Subproblem}

Similarly, we provide a proposition as preparation,  with its proof deferred to the \ref{app:prop3}.
\begin{prop}\label{prop3}
    If $f_4(\mathbf{v})=\mathbf{d}^T\mathbf{v}+\mathbb{I}_{\mathbb{R}_+}(\mathbf{v})$, then for $\tau_4>0$, the closed-form proximal mapping of $f_4$ is given by
    \begin{equation*}
        \mathrm{prox}_{\tau_4f_4}(\mathbf{v})=\max\{\mathbf{v}-\tau\mathbf{d}\}.
    \end{equation*}
\end{prop}

Fixing the dual variable $\lambda^i$, newly updated $\mathbf{w}^{i+1}$, $\mathbf{u}^{i+1}$ and $\mathbf{k}^{i+1}$, the subproblem for $\mathbf{v}$ is
\begin{align}
    \mathbf{v}^{i+1} &=\mathrm{arg}\min_{\mathbf{v}}\mathcal{L}(\mathbf{w}^{i+1}, \mathbf{u}^{i+1}, \mathbf{k}^{i+1}, \mathbf{v}; \lambda^i) \nonumber\\
    &=\mathrm{arg}\min_{\mathbf{v}} f_4(\mathbf{v})+\frac{\rho}{2}\Vert\mathbf{A}[\mathbf{w}^{i+1};\mathbf{k}^{i+1};\mathbf{v};\mathbf{u}^{i+1}]-\frac{\lambda^i}{\rho}\Vert_2^2.
    \label{eq21}
\end{align}
In view of Proposition \ref{prop3}, we linearize the quadratic term to perform one step of the proximal gradient iteration to update $\mathbf{v}$
\begin{align}
    \mathbf{v}^{i+1}=\max\{\tilde{\mathbf{v}}^{i+1}, \mathbf{0}\},
    \label{eq22}
\end{align}
where
\begin{align*}
    \tilde{\mathbf{v}}^{i+1}:=\mathbf{v}^i-\tau_4\mathbf{a}(s^{i+1}+\mathbf{a}^T\mathbf{v})-\tau_4\mathbf{d}
\end{align*}
here, $s^{i+1}=\frac{1}{2}\mathbf{z}^T\mathbf{w}^{i+1}+2\mathbf{b}^T\mathbf{k}^{i+1}-\frac{\lambda_1^i}{\rho}\in\mathbb{R}$ is a constant. 

\subsection{Dual Variables Update}
Subsequently, the dual variables $\lambda$ is updated as:
\begin{align}
    \lambda^{i+1} &= \lambda^i - \rho\mathbf{Ax}^{i+1},
    \label{eq23}
\end{align}
where $\mathbf{x}^{i+1}=[\mathbf{w}^{i+1};\mathbf{k}^{i+1};\mathbf{v}^{i+1};\mathbf{u}^{i+1}]$.

The overall description of our GLOPSS-CS is given in Algorithm \ref{alg1}.
\begin{algorithm}[!ht]
    \caption{Graph Learning from Smooth Signals under Partial Observability with Column Sparsity Regularization (GLOPSS-CS)}
    \label{alg1}
    \renewcommand{\algorithmicrequire}{\textbf{Input:}}
    \renewcommand{\algorithmicensure}{\textbf{Output:}}
    \begin{algorithmic}[1]
        \REQUIRE penalty parameters $\alpha$, $\beta$, $\gamma_{2,1}$ and $\rho$, step sizes $\tau_1$, $\tau_2$, $\tau_3$, $\tau_4$, primal residual tolerance $\epsilon_p$, dual residual tolerance $\epsilon_d$,  
        \ENSURE     $\mathbf{w}^\ast$, $\mathbf{k}^\ast$, $\mathbf{v}^\ast$ and $\mathbf{u}^\ast$   
        \WHILE{$r_p>\epsilon_p$ or $r_d>\epsilon_d$}
           \STATE update $\mathbf{w}$ according to \eqref{eq16}
           \STATE update $\mathbf{u}$ according to \eqref{eq18}
           \STATE update $\mathbf{k}$ according to \eqref{eq20}
           \STATE update $\mathbf{v}$ according to \eqref{eq22}
           \STATE update $\lambda$ according to \eqref{eq23}
           \STATE set primal residual $r_p=\Vert \mathbf{Bw}^{i+1}-\mathbf{u}^{i+1}\Vert_2$
           \STATE set dual residual $r_d=\Vert\rho_2\mathbf{B}^T(\mathbf{u}^{i+1}-\mathbf{u}^i)\Vert_2$
           \STATE $i\leftarrow i+1$
        \ENDWHILE
    \end{algorithmic}
\end{algorithm}

\subsection{Low Rank Regularization}
We have discussed the column-sparsity pattern of $\tilde{\mathbf{K}}$ above, while setting $\gamma_\ast=0$. And next, we consider the nuclear norm regularization of $\tilde{\mathbf{K}}$ to promote a solution with low rank on $\tilde{\mathbf{K}}$. Similarly, the algorithm (GLOPSS-LR) is the same as before, except the $\mathbf{k}$-subproblem. When $\gamma_{2,1}=0$, we have the following lemma adapted from \cite{beck2017first}.
\begin{lem}\label{lemma3}
     For a matrix $\mathbf{Y}\in\mathbb{R}^{m\times n}$, consider:
    \begin{equation*}
        \min_{\mathbf{X}\in\mathbb{R}^{m\times n}} \nu \Vert\mathbf{X}\Vert_\ast + \frac{1}{2}\Vert\mathbf{X-Y}\Vert_F^2.
    \end{equation*}
    The optimal solution is:
    \begin{equation*}
        \mathbf{X}:=S(\mathbf{Y},\nu)=\mathbf{U}\mathrm{Diag}(s(\sigma,\nu))\mathbf{V}^T,
    \end{equation*}
    where the singular value decomposition of $\mathbf{Y}$ is $\mathbf{Y}=\mathbf{U}\mathrm{Diag}(\sigma)\mathbf{V}^T$, and the thresholding operator is $s(\mathbf{x},\nu):=\bar{\mathbf{x}}$ with
    \begin{equation*}
        \bar{x}_i:=\begin{cases}
        x_i - \nu,& \text{if $x_i-\nu>0$}\\
        0,& \text{otherwise.}
        \end{cases}
    \end{equation*}
\end{lem}

Now, the $\mathbf{k}$-subproblem becomes
\begin{align}
    \mathbf{k}^{i+1}&=\mathrm{arg}\min_{\mathbf{k}\in\mathbb{R}^{o^2}}\mathcal{L}(\mathbf{w}^{i+1}, \mathbf{u}^{i+1}, \mathbf{k}, \mathbf{v}^i; \lambda_1^i, \lambda_2^i) \nonumber\\
    &=2\mathbf{b}^T\mathbf{k}+\gamma_\ast\Vert\tilde{\mathbf{K}}\Vert_\ast +\frac{\rho}{2}(\frac{1}{2}\mathbf{z}^T\mathbf{w}^{i+1}+2\mathbf{b}^T\mathbf{k}+\mathbf{a}^T\mathbf{v}^{i}-\frac{\lambda_1^{i}}{\rho}) \nonumber\\
    &= f_{31}(\mathbf{k})+f_{32}(\mathbf{k}),
    \label{eq24}
\end{align}
where,
\begin{align*}
    f_{31}(\mathbf{k})&:=2\mathbf{b}^T\mathbf{k}+\frac{\rho}{2}(\frac{1}{2}\mathbf{z}^T\mathbf{w}^{i+1}+2\mathbf{b}^T\mathbf{k}+\mathbf{a}^T\mathbf{v}^{i}-\frac{\lambda_1^{i}}{\rho})^2,\quad f_{32}(\mathbf{k}) :=\gamma_\ast\Vert\tilde{\mathbf{K}}\Vert_\ast.
\end{align*}
 In view of Lemma \ref{lemma3}, we linearize the quadratic term in $f_{31}(\mathbf{k})$ to perform one step of the proximal gradient iteration to update $\mathbf{k}$
\begin{align}
    \mathbf{k}^{i+1} & = \mathrm{prox}_{\tau_3f_{32}}(\mathbf{k}^i-\tau_32(1+\frac{1}{2}\rho\mathbf{z}^T\mathbf{w}^{i+1}+2\rho\mathbf{b}^T\mathbf{k}^i+\rho\mathbf{a}^T\mathbf{v}^{i}-\lambda_1^i)\mathbf{b}) \nonumber\\
    &= \text{vec}(\mathbf{U}^i\mathrm{Diag}(s(\sigma^i,\gamma_\ast))(\mathbf{V}^i)^T),
    \label{eq25}
\end{align}
where $\mathbf{U}^i$ and $\mathbf{V}^i$ are composed by the left and right singular vectors of $\text{vec}^{-1}(\tilde{\mathbf{k}}^{i+1})$, i.e., $\text{vec}^{-1}(\tilde{\mathbf{k}}^{i+1})=\mathbf{U}^i\mathrm{Diag}(\sigma^i)(\mathbf{V}^i)^T$. Here we have $\tilde{\mathbf{k}}^{i+1}=\mathbf{k}^i-\tau_32(1+\frac{1}{2}\rho\mathbf{z}^T\mathbf{w}^{i+1}+2\rho\mathbf{b}^T\mathbf{k}^i+\rho\mathbf{a}^T\mathbf{v}^{i}-\lambda_1^i)\mathbf{b}.$ And $s(\sigma^i,\gamma_\ast)$ is the thresholding operator: 
\begin{equation*}
        s(\sigma^i,\gamma_\ast)=\bar{\sigma}^i, \mathrm{with}\quad \bar{\sigma}^i_t:=\begin{cases}
\sigma^i_t - \nu,& \text{if $x_i-\nu>0$}\\
0,& \text{otherwise}
\end{cases}
\end{equation*}
where $t\in\{1,2,\ldots, o\}$.

The overall description of our GLOPSS-LR is given in Algorithm \ref{alg2}.
\begin{algorithm}[!ht]
    \caption{Graph Learning from Smooth Signals under Partial Observability with Low-Rank Regularization (GLOPSS-LR)}
    \label{alg2}
    \renewcommand{\algorithmicrequire}{\textbf{Input:}}
    \renewcommand{\algorithmicensure}{\textbf{Output:}}
    \begin{algorithmic}[1]
        \REQUIRE penalty parameters $\alpha$, $\beta$, $\gamma_{\ast}$, $\rho$, step sizes $\tau_1$, $\tau_2$, $\tau_3$, $\tau_4$, primal residual tolerance $\epsilon_p$, dual residual tolerance $\epsilon_d$,  
        \ENSURE     $\mathbf{w}^\ast$, $\mathbf{k}^\ast$, $\mathbf{v}^\ast$ and $\mathbf{u}^\ast$    
        \WHILE{$r_p>\epsilon_p$ or $r_d>\epsilon_d$}
           \STATE update $\mathbf{w}$ according to \eqref{eq16}
           \STATE update $\mathbf{u}$ according to \eqref{eq18}
           \STATE update $\mathbf{k}$ according to \eqref{eq25}
           \STATE update $\mathbf{v}$ according to \eqref{eq22}
           \STATE update $\lambda$ according to \eqref{eq23}
           \STATE set primal residual $r_p=\Vert \mathbf{Bw}^{i+1}-\mathbf{u}^{i+1}\Vert_2$
           \STATE set dual residual $r_d=\Vert\rho_2\mathbf{B}^T(\mathbf{u}^{i+1}-\mathbf{u}^i)\Vert_2$
           \STATE $i\leftarrow i+1$
        \ENDWHILE
    \end{algorithmic}
\end{algorithm}

\subsection{Computational Complexity}

The update step in \eqref{eq16} exhibits a computational complexity of $\mathcal{O}(o^2)$. Furthermore, given $\mathbf{w}$, $\mathbf{k}$, and $\mathbf{v}$, the update for $\mathbf{u}$ in \eqref{eq18} also requires $\mathcal{O}(o^2)$ operations. Similarly, computing $\mathbf{k}$ in \eqref{eq20} and performing the update in \eqref{eq22} each demand $\mathcal{O}(o^2)$ operations. Thus, every iteration of Algorithm \ref{alg1} has an overall complexity of $\mathcal{O}(o^2)$.

In contrast, the primary distinction in Algorithm \ref{alg2} lies in the update of $\mathbf{k}$. Specifically, as shown in \eqref{eq25}, the update involves a singular value decomposition (SVD) of a matrix, incurring a computational cost of $\mathcal{O}(o^3)$ per iteration. Moreover, when an off-the-shelf solver (e.g., CVX) is utilized, the complexity escalates to $\mathcal{O}(o^7)$. Therefore, compared to traditional methods, the proposed algorithms offer reduced computational complexity, which promotes faster convergence and renders them well-suited for large-scale graph learning scenarios.
\section{Convergence Analysis}\label{sect4}
Let the positive constants $\alpha, \beta$, and $\gamma_{2,1} (\gamma_\ast)$ be given. As noted earlier, Problem \eqref{eq11} has a unique globally optimal minimizer $(\mathbf{w}^\ast, \mathbf{u}^\ast, \mathbf{k}^\ast, \mathbf{v}^\ast)$. In this section, we prove that both Algorithm \ref{alg1} (GLOPSS-CS) and \ref{alg2} (GLOPSS-LR) converge globally, following the approach of Hong et al.\cite{hong2017linear} and Ma et al.\cite{ma2016alternating}.

Since \eqref{eq11} is a  convex program with linear constraints and at least one solution, its Karush-Kuhn-Tucker (KKT) conditions are necessary and sufficient for optimality. Denoting the dual multipliers by $\lambda_1$, $\lambda_2$, the KKT system can be written compactly as
\begin{align}
    \mathbf{0} &\in \partial f_1(\mathbf{w}) - \frac{1}{2}\mathbf{z}\lambda_1-\mathbf{B}^T\lambda_2, \nonumber\\
    \mathbf{0} &\in \partial f_2(\mathbf{u})+\lambda_2, \nonumber\\
    \mathbf{0} &\in \partial f_3(\mathbf{k})-2\mathbf{b}\lambda_1, \nonumber\\
    \mathbf{0} &\in \partial f_4(\mathbf{v})-\mathbf{a}\lambda_1, \nonumber\\
    \mathbf{0} &= \frac{1}{2}\mathbf{z}^T\mathbf{w}+2\mathbf{b}^T\mathbf{k}+\mathbf{a}^T\mathbf{v}, \nonumber\\
    \mathbf{0} &= \mathbf{Bw}-\mathbf{u}.
    \label{eq26}
\end{align}
Let $\mathbb{Q}^\ast$ be the set of all points  $[\mathbf{w};\mathbf{u};\mathbf{k};\mathbf{v};\lambda_1;\lambda_2]$ satisfying these conditions. Since each iteration of GLOPSS-CS and GLOPSS-LR updates both primal and dual variables, it is natural to investigate whether the sequence $\{[\mathbf{w}^i;\mathbf{u}^i;\mathbf{k}^i;\mathbf{v}^i;\lambda_1^i;\lambda_2^i]\}_{i\geq 0}$ converges to a member of $\mathbb{Q}^\ast$, and at what rate. 

\subsection{Global Linear Convergence}
Our casting GLOPSS within the ADMM framework studied in \cite{hong2017linear}, we can directly invoke their results to assert global linear convergence, with its proof deferred to the \ref{app:thm_1}.
\begin{thm}\label{thm_1}
    Suppose the step sizes $\tau_1$, $\tau_2$, $\tau_3$ and $\tau_4$ are chosen sufficiently small. Then the sequence of iterations $\{\mathbf{w}^i, \mathbf{u}^i, \mathbf{k}^i, \mathbf{v}^i;\lambda^i\}$ generated by the GLOPSS-CS and GLOPSS-LR algorithms converge linearly to some KKT point $[\mathbf{w}^\ast;\mathbf{u}^\ast;\mathbf{k}^\ast;\mathbf{v}^\ast;\lambda^\ast]\in\mathbb{Q}^\ast$.
\end{thm}

Theorem \ref{thm_1} reveals that the iteration sequence generated by GLOPSS-LR and GLOPSS-CS converge to some point in $\mathbb{Q}^\ast$ at the \textit{linear} rate. To the best of our knowledge, Theorem \ref{thm_1} furnishes the convergence guarantee known to date for addressing the graph learning formulation \eqref{eq11}.

Moreover, based on the structure of GLOPSS-CS and GLOPSS-LR, we have found a connection between the step sizes and the maximum singular value of the coefficient matrices of the linear constraints in \eqref{eq11}. Therefore, in the next subsection, we provide an upper bound on the step size that still ensures linear convergence, to guide the selection of step sizes in practical graph learning.

\subsection{Selection of Step-sizes}
To make the above result constructive, we relate allowable step-sizes to the largest singular values of the matrices defining the linear constraints. Following the technique in \cite{ma2016alternating},  we commence by stating the following theorem, with its proof deferred to the \ref{app:thm_2}:
\begin{thm}\label{thm_2}
    Suppose that $(\mathbf{w}^\ast, \mathbf{u}^\ast, \mathbf{k}^\ast, \mathbf{v}^\ast)$ is the optimal solution of Problem \eqref{eq11} and $\lambda^\ast$ are the corresponding optimal dual variables. If the step sizes adhere to $\tau_1<\frac{1}{\sigma_{\max}^2(\mathbf{M}_1)}$, $\tau_2<\frac{1}{\sigma_{\max}^2(\mathbf{M}_2)}$, $\tau_3<\frac{1}{\sigma_{\max}^2(\mathbf{M}_3)}$, $\tau_4<\frac{1}{\sigma_{\max}^2(\mathbf{M}_4)}$, then there exists
    \begin{align*}
        c = \min\{\frac{\rho}{\tau_1}-\rho\sigma_{\max}^2(\mathbf{M}_1)), \frac{\rho}{\tau_2}-\rho\sigma_{\max}^2(\mathbf{M}_2)), \frac{\rho}{\tau_3}-\rho\sigma_{\max}^2(\mathbf{M}_3)), \frac{\rho}{\tau_4}-\rho\sigma_{\max}^2(\mathbf{M}_4)), \frac{1}{\rho}-\mu\}>0
    \end{align*} 
    such that the sequence $\{(\mathbf{w}^i,\mathbf{u}^i, \mathbf{k}^i, \mathbf{v}^i, \lambda^i)\}_{i=0}^\infty$ generated by Algorithm \ref{alg1} and \ref{alg2} satisfies
    \begin{equation*}
        \Vert\mathbf{y}^i-\mathbf{y}^\ast\Vert_\mathbf{M}^2-\Vert\mathbf{y}^{i+1}-\mathbf{y}^\ast\Vert_\mathbf{M}^2 \geq c \Vert\mathbf{y}^i-\mathbf{y}^{i+1}\Vert_\mathbf{M}^2,
    \end{equation*}
    where $\sigma_{\max}(\cdot)$ denotes the largest singular value, $\mathbf{y}^i=[\mathbf{w}^i;\mathbf{u}^i;\mathbf{k}^i;\mathbf{v}^i;\lambda^i]$, $\mathbf{y}^\ast=[\mathbf{w}^\ast; \mathbf{u}^\ast;\mathbf{k}^\ast;\mathbf{v}^\ast;\lambda^\ast]$, $\mathbf{M}=\text{Diag}[\frac{\rho}{\tau_1}\mathbf{I}-\rho\mathbf{M}_1^T\mathbf{M}_1;\frac{\rho}{\tau_2}\mathbf{I}-\rho\mathbf{M}_2^T\mathbf{M}_2;\frac{\rho}{\tau_3}\mathbf{I}-\rho\mathbf{M}_3^T\mathbf{M}_3;\frac{\rho} {\tau_4}\mathbf{I};\frac{1}{\rho}\mathbf{I}]$, and $\Vert\mathbf{y}\Vert_\mathbf{M}=\sqrt{\mathbf{y}^T\mathbf{My}}$ for $\mathbf{y}\in\mathbb{R}^{o^2+2o+p+3}$. Here, $\mu=(\frac{1}{2}+\frac{\tau_4\sigma_{\max}^2(\mathbf{M}_4)}{2})/\rho$.
\end{thm}

As Theorem \ref{thm_2} suggests, the largest singular values of $\mathbf{M}_1$, $\mathbf{M}_2$, $\mathbf{M}_3$ and $\mathbf{M}_4$ dictate the maximum allowable values of $\tau_1$, $\tau_2$, $\tau_3$ and $\tau_4$ that ensure the global convergence of Algorithm \ref{alg1} and \ref{alg2}. To guide the selection of step sizes, we present the following proposition, with its proof deferred to the \ref{app:prop4}:
\begin{prop}\label{prop4}
    The spectral norms of $\mathbf{M}_1$, $\mathbf{M}_2$, $\mathbf{M}_3$ and $\mathbf{M}_4$ satisfy\\
    $(1)$ $\Vert\mathbf{M}_1\Vert_2\leq\frac{1}{2}\kappa o+\sqrt{2(o-1)}$; $(2)$ $\Vert\mathbf{M}_2\Vert_2=1$; $(3)$ $\Vert\mathbf{M}_3\Vert_2=2\sqrt{o}$; $(4)$ $\Vert\mathbf{M}_4\Vert=2$, where $\kappa^2$ is the maximal entry of $\mathbf{zz}^T$.
\end{prop}

Define the projection-based KKT residual mapping $\mathcal{E}(\mathbf{y})$ by
\begin{align}
    \mathcal{E}(\mathbf{w},\mathbf{u},\mathbf{k},\mathbf{v},\lambda):=\begin{bmatrix}
        \mathbf{w}-P_{X_1}(\mathbf{w}-(\partial f_1(\mathbf{w})-\mathbf{M}_1^T\lambda))\\
        \mathbf{u}-P_{X_2}(\mathbf{u}-(\partial f_2(\mathbf{u})-\mathbf{M}_2^T\lambda))\\
        \mathbf{k}-P_{X_3}(\mathbf{k}-(\partial f_3(\mathbf{k})-\mathbf{M}_3^T\lambda))\\
        \mathbf{v}-P_{X_4}(\mathbf{v}-(\partial f_4(\mathbf{v})-\mathbf{M}_4^T\lambda))\\
        \mathbf{Ax}
    \end{bmatrix},
    \label{eq28}
\end{align}
where $P_{\mathcal{C}}(x):=\text{arg}\min_{y\in \mathcal{C}}\Vert x-y\Vert_2$ denotes the projection operator onto the feasible set $\mathcal{C}$. $\mathbf{X}=X_1\times X_2\times X_3\times X_4$ denotes the feasible set with $\mathbf{w}\in X_1$, $\mathbf{u}\in X_2$, $\mathbf{k}\in X_3$, and $\mathbf{v}\in X_4$.

In particular, the following result, which shows that the mapping $\mathcal{E}(\mathbf{y})$ possesses the so-called error bound property, is a direct consequence of \cite{wang2023linearly}:
\begin{lem}\label{lemma4}
    The set-valued mapping $\mathcal{E}(\mathbf{y})$ is metrically subregular at any KKT point of Problem \eqref{eq11}. Concretely, there exists a positive constant $\zeta$ and a open neighborhood $\mathcal{U}\subseteq\mathbb{R}^{\ell}$ (where $\ell=o^2+2o+p+3$) containing the solution set $\mathbb{Q}^\ast$, such that for every point $\mathbf{y}\in\mathcal{U}$, the following inequality  holds:
    \begin{equation*}
        \text{dist}(\mathbf{y},\mathbb{Q}^\ast)\leq\zeta\text{dist}(\mathbf{0},\mathcal{E}(\mathbf{y})).
    \end{equation*}
\end{lem}

\begin{remark}
    In other words, once $\mathbf{y}$ lies sufficiently close to a KKT point, its distance to the exact solution set can be bounded above (up to the scaler factor $\zeta$) by the norm of the residual $\mathcal{E}(\mathbf{y})$. This property underpins both local error-estimation and convergence-rate analyses for algorithms solving Problem \eqref{eq11}.
\end{remark}

To get the linear convergence of GLOPSS-CS and GLOPSS-LR, we need to establish relationship between $\Vert\mathbf{y}^{i+1}-\mathbf{y}^i\Vert_{\mathbf{M}}$ and $\text{dist}(\mathbf{0},\mathcal{E}(\mathbf{y}^{i+1})$, which is shown in the following lemma with its proof deferred to the \ref{app:lemma5}:
\begin{lem}\label{lemma5}
    Let $\mathbf{y}^{i+1}=\{\mathbf{w}^i,\mathbf{u}^i,\mathbf{k}^i,\mathbf{v}^i,\lambda^i\}$ be the sequence generated by GLOPSS-LR or GLOPSS-CS. There exists $\eta>0$ such that
    \begin{equation*}
        \Vert\mathbf{y}^i-\mathbf{y}^{i+1}\Vert_{\mathbf{M}}^2\geq\eta\text{dist}^2(\mathbf{0},\mathcal{E}(\mathbf{y}^{i+1})).
    \end{equation*}
\end{lem}
Combining Lemma \ref{lemma5} and the inequality $\text{dist}(\mathbf{y},\mathbb{Q}^\ast)\leq\zeta\text{dist}(\mathbf{0},\mathcal{E}(\mathbf{y}))$ in Lemma \ref{lemma4}, we have
\begin{align}
    \Vert\mathbf{y}^{i+1}-\mathbf{y}^i\Vert_{\mathbf{M}}^2\geq\eta\text{dist}^2(\mathbf{0},\mathcal{E}(\mathbf{y}^{i+1}))\geq\frac{\eta}{\zeta^2}\text{dist}^2(\mathbf{y}^{i+1},\mathbb{Q}^\ast).
    \label{eq29}
\end{align}
Select $\mathbf{y}^\ast\in\mathbb{Q}^\ast$ such that $\text{dist}_{\mathbf{M}}(\mathbf{y}^{i},\mathbb{Q}^\ast)=\Vert\mathbf{y}^i-\mathbf{y}^\ast\Vert_{\mathbf{M}}$. Then,  we have
\begin{align}
    (1+\frac{c\eta}{\zeta^2})\text{dist}^2_{\mathbf{M}}(\mathbf{y}^{i+1},\mathbb{Q}^\ast)&\leq\Vert\mathbf{y}^{i+1}-\mathbf{y}^\ast\Vert_\mathbf{M}^2+\frac{c\eta}{\zeta^2}\text{dist}^2_{\mathbf{M}}(\mathbf{y}^{i+1},\mathbb{Q}^\ast) \nonumber\\
    &\leq \Vert\mathbf{y}^{i+1}-\mathbf{y}^\ast\Vert_\mathbf{M}^2+c\Vert\mathbf{y}^{i+1}-\mathbf{y}^i\Vert_\mathbf{M}^2 \nonumber\\
    &\leq\Vert\mathbf{y}^i-\mathbf{y}^\ast\Vert_\mathbf{M}^2 \nonumber\\
    &= \text{dist}^2_{\mathbf{M}}(\mathbf{y}^i,\mathbb{Q}^\ast),
    \label{eq30}
\end{align}
where the second and the third inequalities follow from \eqref{eq29} and Theorem \ref{thm_2}. Now we draw the linear convergence of GLOPSS-CS and GLOPSS-LR in the following theorem.
\begin{thm}\label{thm_3}
    Suppose the step sizes adhere to $\tau_1<\frac{1}{\sigma_{\max}^2(\mathbf{M}_1)}$, $\tau_2<\frac{1}{\sigma_{\max}^2(\mathbf{M}_2)}$, $\tau_3<\frac{1}{\sigma_{\max}^2(\mathbf{M}_3)}$, $\tau_4<\frac{1}{\sigma_{\max}^2(\mathbf{M}_4)}$, let $\{\mathbf{w}^i, \mathbf{u}^i,\mathbf{k}^i,\mathbf{v}^i,\lambda^i\}$ be the sequence generated by Algorithm \ref{alg1} or \ref{alg2}. Then, there exists $0<\varrho<1$ such that
    \begin{equation*}
        \text{dist}_{\mathbf{M}}(\mathbf{y}^{i+1},\mathbb{Q}^\ast) \leq \varrho \text{dist}_{\mathbf{M}}(\mathbf{y}^i,\mathbb{Q}^\ast).
    \end{equation*}
\end{thm}

\begin{remark}
    Theorem \ref{thm_3} ensures the linear convergence of Algorithm \ref{alg1} and \ref{alg2} with an arbitrary initial point, provided that appropriate step sizes are chosen. In light of Proposition \ref{prop4}, we should select $\tau_1<\frac{1}{(\frac{1}{2}\kappa o+\sqrt{2(o-1)})^2}$, $\tau_2<1$, $\tau_3<\frac{1}{4o}$ and $\tau_4<\frac{1}{2}$ when addressing the graph learning model. This convergence analysis not only validates the efficacy of our GLOPSS-CS and GLOPSS-LR algorithms but also provides practical guidelines for selecting step sizes, thereby enhancing its applicability in solving graph learning problems.
\end{remark}

In summary, our analysis not only guarantees that GLOPSS-CS and GLOPSS-LR find a KKT point of \eqref{eq11} from any initialization, but also prescribes explicit stepsizes to achieve global linear convergence. To the best of our knowledge, our work represents the pioneering development of a first-order method tailored to the graph learning formulation encapsulated in \eqref{eq11}, accompanied by a linear convergence guarantee.

\subsection{Adaptive Penalty for Partial Observability Problem}
To extend our ADMM-based topology inference framework to networks containing unobserved entities, we first introduce an adaptive penalty mechanism that ensures robust convergence despite hidden nodes. We then establish precise recovery guarantees quantifying the impact of unobserved nodes on estimation accuracy.

In what follows, we formalize the adaptive penalty strategy and prove its linear convergence under a mild identifiability condition, with its proof deferred to the \ref{app:thm5}.
\begin{definition}[Hidden Node Identifiability]
A network with hidden nodes is said to satisfy the identifiability condition with parameter $\delta > 0$ if:
\begin{equation}
\sigma_{\min}\left(\sum_{i=1}^n \mathbf{x}_i^{(o)} (\mathbf{x}_i^{(o)})^T \otimes \mathbf{I}_m\right) \geq \delta n
\end{equation}
where $\mathbf{x}_i^{(o)} \in \mathbb{R}^o$ is the signal vector of observed nodes at time $i$.
\end{definition}

\begin{thm}
\label{thm:hidden_node_convergence}
Consider the network topology inference problem with hidden nodes, where we observe signals $\mathbf{X}_{\mathcal{O}} \in \mathbb{R}^{o \times n}$ from $o$ observed nodes over $n$ time instances. Let the GLOPSS algorithm use the adaptive penalty parameter update:
\begin{equation}
\rho^{i+1} = \begin{cases}
\tau_{\text{inc}} \rho^i & \mathrm{if } \|\mathbf{r}_p^{i}\|_2 > \mu \|\mathbf{r}_d^{i}\|_2 \\
\rho^i / \tau_{\text{dec}} & \mathrm{if } \|\mathbf{r}_d^{i}\|_2 > \mu \|\mathbf{r}_p^{i}\|_2 \\
\rho^i & \mathrm{otherwise}
\end{cases}
\end{equation}
where $\mathbf{r}_p^{i}$ and $\mathbf{r}_d^{i}$ are the primal and dual residuals in the augmented system including hidden node variables.

Define the observability coefficient as $\xi := \frac{o}{m} = \frac{o}{o+h}$ and assume the network satisfies the hidden node identifiability condition with parameter $\delta > 0$. Then the algorithm converges linearly with rate
\begin{equation}
\|\mathbf{y}^i - \mathbf{y}^*\|_2 \leq \left(1 - \frac{\xi \delta \sigma_{\min}(\mathbf{A}^T\mathbf{A})}{C \max\{\rho_{\max}, L_{\max}\}}\right)^i \|\mathbf{y}_0 - \mathbf{y}^*\|_2
\end{equation}
where $C > 0$ is a constant, $\rho_{\max} = \sup_{k \geq 0} \rho_k$, and $L_{\max} = \max\{2\beta, \alpha/\min_j u_j^2\}$.
\end{thm}

\begin{remark}
The convergence rate in Theorem \ref{thm:hidden_node_convergence} exhibits two critical dependencies that characterize the fundamental trade-offs in network topology inference with hidden nodes: The linear convergence rate scales directly with the observability coefficient $\xi = \frac{o}{m}$, indicating that networks with fewer hidden nodes ($h$ small, hence $\xi$ close to 1) achieve faster convergence. This reflects the intuitive principle that increased observability enhances algorithmic performance. The parameter $\delta$ from the hidden node identifiability condition provides a lower bound on convergence speed. Networks satisfying stronger identifiability conditions (larger $\delta$) exhibit superior convergence rates, emphasizing the importance of structural assumptions in computational tractability.

The multiplicative relationship $\xi \delta$ in the convergence factor reveals that both observability and identifiability must be jointly favorable for optimal performance. Notably, when $\xi \to 0$ (predominantly hidden network) or $\delta \to 0$ (weak identifiability), the convergence rate degrades significantly, potentially approaching the non-convergent regime. This analysis provides theoretical guidance for network design and algorithm parameter selection in practical applications.
\end{remark}

\begin{remark}
The convergence rate in Theorem~\ref{thm:hidden_node_convergence} validates assumptions \textbf{(A1)} and \textbf{(A2)}: the linear dependence on observability coefficient $\xi = \frac{o}{m}$ demonstrates that $h \ll o$ is essential for algorithmic efficiency, while the identifiability parameter $\delta$ reflects the signal smoothness requirement. This $\xi$-$\delta$ interplay establishes a fundamental trade-off between network observability and signal regularity, providing theoretical justification for the problem formulation.
\end{remark}

Building upon this convergence analysis, we now quantify the error in recovering the observed‐node subgraph under noisy measurements.

\subsection{Recovery Guarantee with Hidden Nodes}
We have the following recovery error bound, with its proof deferred to the \ref{app:thm6}:
\begin{thm}
\label{thm:hidden_recovery}
Consider the network with $o$ observed nodes and $h$ hidden nodes under noisy observations $\mathbf{X}_{\mathcal{O}} = \mathbf{X}_{\mathcal{O}}^* + \mathbf{E}$, where $\mathbf{E}_{ij} \sim \mathcal{N}(0, \sigma^2)$. Let $\hat{\mathbf{L}}_{\mathcal{O}}$ be the recovered submatrix of connections among observed nodes. Then, with probability at least $1-\theta$,
\begin{equation}
\|\hat{\mathbf{L}}_{\mathcal{O}} - \mathbf{L}_{\mathcal{O}}^{\text{eff}}\|_F \leq C\sigma\sqrt{\frac{s_o \log(o/\theta)}{n}} \cdot \left(1 + \frac{h}{o}\right)
\end{equation}
where $\mathbf{L}_{\mathcal{O}}^{\text{eff}} = \mathbf{L}_{\mathcal{O}}^* - \mathbf{L}_{\mathcal{OH}}^* (\mathbf{L}_{\mathcal{HH}}^*)^{-1} \mathbf{L}_{\mathcal{HO}}^*$ is the effective Laplacian among observed nodes, $s_o$ is the sparsity of $\mathbf{L}_{\mathcal{O}}^{\text{eff}}$, i.e., $s_o=|\{(i,j)\mid [\mathbf{L}_{\mathcal{O}}^{\text{eff}}]_{ij}\neq 0\}|$, and $C > 0$ is a universal constant.
\end{thm}

\begin{remark}
The bound in Theorem~\ref{thm:hidden_recovery} reveals several key insights about network recovery in the presence of hidden nodes: The recovery algorithm estimates $\mathbf{L}_{\mathcal{O}}^{\text{eff}}$ rather than the true $\mathbf{L}_{\mathcal{O}}^*$, which accounts for the marginal effects of hidden nodes through the Schur complement $\mathbf{L}_{\mathcal{OH}}^* (\mathbf{L}_{\mathcal{HH}}^*)^{-1} \mathbf{L}_{\mathcal{HO}}^*$. The factor $(1 + h/o)$ quantifies the degradation due to unobserved nodes, showing that the recovery error scales linearly with the ratio of hidden to observed nodes. The bound exhibits the standard $\mathcal{O}(\sqrt{s_o \log o / n})$ rate for sparse recovery, where the effective sparsity $s_o$ of $\mathbf{L}_{\mathcal{O}}^{\text{eff}}$ may differ from that of $\mathbf{L}_{\mathcal{O}}^*$ due to hidden node interactions. The linear dependence on noise level $\sigma$ is optimal for Gaussian noise, matching minimax lower bounds in sparse estimation.
\end{remark}

\section{Numerical Experiments}\label{sect5}
This section presents numerical experiments to evaluate and compare the proposed algorithms. We assess their performance using synthetic data and real-world datasets, implemented in MATLAB. For comparison, because the classic ADMM framework only has two block, we may consider grouping the $4$ blocks in \eqref{eq11} into $2$ blocks. Here, we group $\mathbf{w}$ and $\mathbf{v}$ as one block and $\mathbf{u}$ and $\mathbf{k}$ as another one\footnote{Theoretically, one can group the $4$ block variables into two blocks with any combination. However, the numerical performance of  different partitions of the variables can vary significantly, as the step sizes of the proximal gradient steps are linked to the largest singular values of the coefficient matrices. The partition in \eqref{equ27} allows larger step size for the first subproblem, thus, GraSS is faster than any other grouping variables methods except the no grouping one.}, therefore, the constraint in \eqref{eq11} can be rewritten as
\begin{align}
    \begin{pmatrix}
        \frac{1}{2}\mathbf{z}^T & \mathbf{a}^T\\
        \mathbf{B} & \mathbf{0}
    \end{pmatrix}\begin{pmatrix}
        \mathbf{w}\\
        \mathbf{v}
    \end{pmatrix}+\begin{pmatrix}
       2\mathbf{b}^T & \mathbf{0}\\
        \mathbf{0} & -\mathbf{I}
    \end{pmatrix}\begin{pmatrix}
        \mathbf{k}\\
        \mathbf{u}
    \end{pmatrix}=\mathbf{0}.
    \label{equ27}
\end{align}
For convenience in describing the experiment later, we will refer the variables grouping method discussed above as GraSS. Similar to GLOPSS, GraSS has two variants based on different regularization constraints, namely GraSS-CS and GraSS-LR. Based on the numerical experiments, we have discovered that GraSS-LR and GraSS-CS, with smaller maximum singular values of the coefficient matrices among all possible variable partitions, are expected to perform faster than other partitioning methods. Given the close performance between GraSS-CS and GraSS-LR, we elect to compare only one of them (GraSS-LR) with GLOPSS-LR, which does not consider variable partitioning. We will focus on comparing the convergence performance of GLOPSS-LR and GraSS-LR. We will also show the runtime comparison of GLOPSS and Gsm-GL (Gsm-LR) proposed by \cite{buciulea2022learning}, who use CVX as a solver. After that, we will also compare GLOPSS with GL-SigRep, as described in \cite{dong2016learning}, highlighting the advantage of considering hidden nodes in  the recovery process. The parameters $\alpha$, $\beta$ and $\gamma_{2,1}$(or $\gamma_\ast$) in Problem \eqref{eq11} are best-tuned to maximize the quality of the learned graphs based on the F-measure. Additionally, parameters $\rho$ and step size $\tau$ in GraSS-LR and GLOPSS-LR with GLOPSS-CS are also finely-tuned to achieve optimal convergence rates.
\begin{figure*}[!htb]
	\centering
	\begin{subfigure}{0.3\linewidth}
		\centering
		\includegraphics[width=0.9\linewidth]{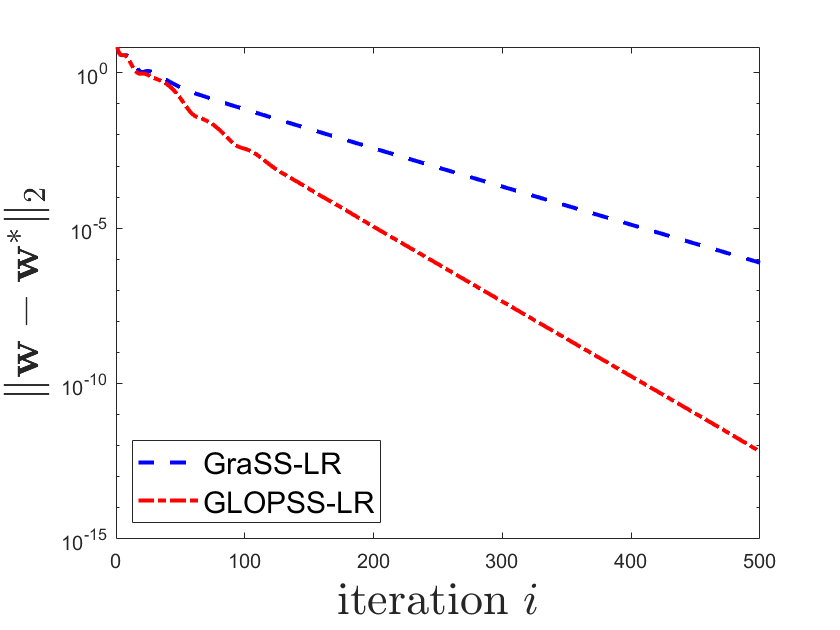}
		\caption{GA graph ($o=20$, $h=1$, $n=100$)}
		\label{G20}
	\end{subfigure}
	\centering
	\begin{subfigure}{0.3\linewidth}
		\centering
		\includegraphics[width=0.9\linewidth]{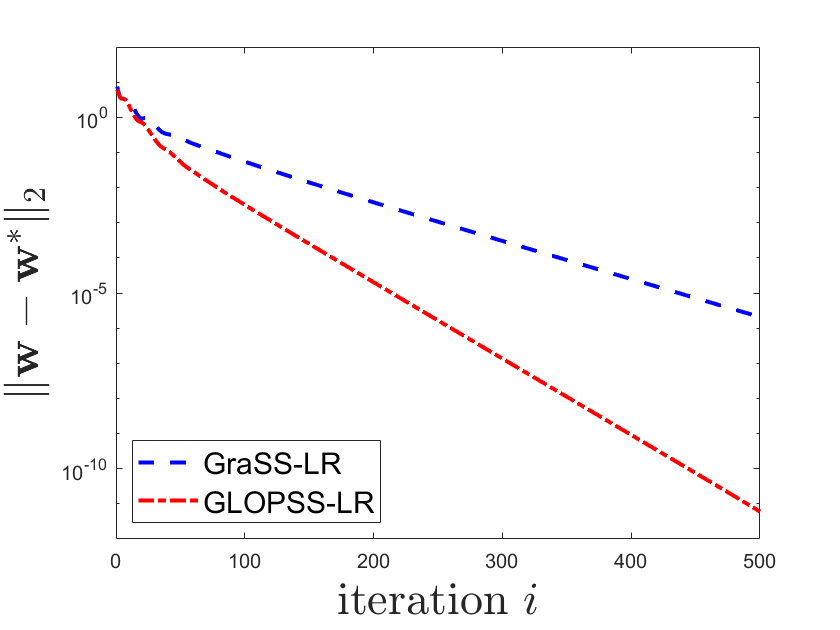}
		\caption{ER graph ($o=20$, $h=1$, $n=100$)}
		\label{E20}
	\end{subfigure}
	\centering
	\begin{subfigure}{0.3\linewidth}
		\centering
		\includegraphics[width=0.9\linewidth]{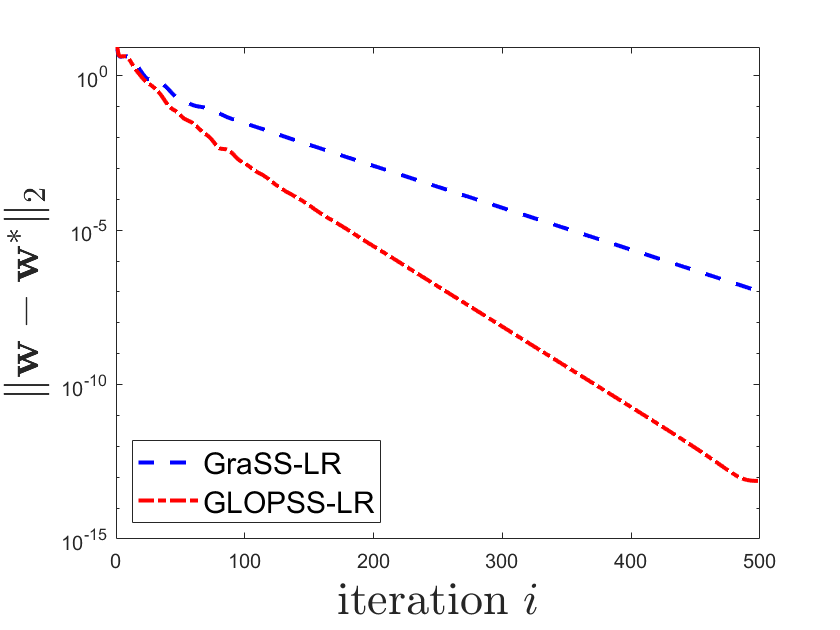}
		\caption{PA graph ($o=20$, $h=1$, $n=100$)}
		\label{P20}
	\end{subfigure}

    \begin{subfigure}{0.3\linewidth}
		\centering
		\includegraphics[width=0.9\linewidth]{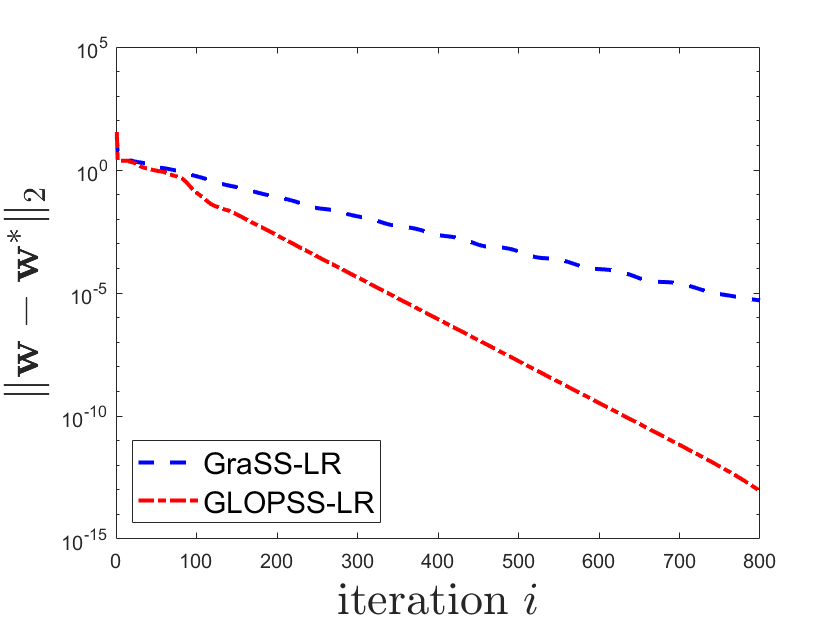}
		\caption{GA graph ($o=50$, $h=1$, $n=400$)}
		\label{G21}
	\end{subfigure}
	\centering
	\begin{subfigure}{0.3\linewidth}
		\centering
		\includegraphics[width=0.9\linewidth]{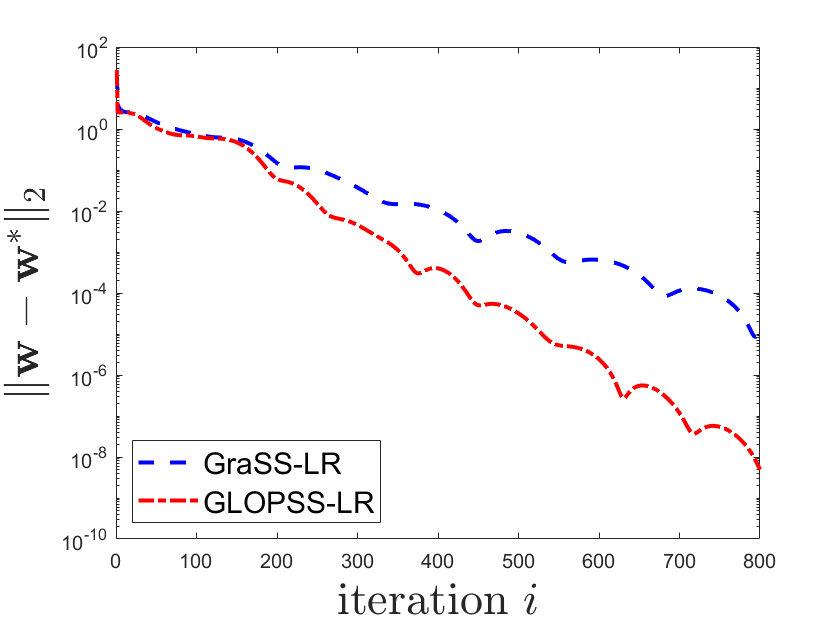}
		\caption{ER graph ($o=50$, $h=1$, $n=400$)}
		\label{E21}
	\end{subfigure}
	\centering
	\begin{subfigure}{0.3\linewidth}
		\centering
		\includegraphics[width=0.9\linewidth]{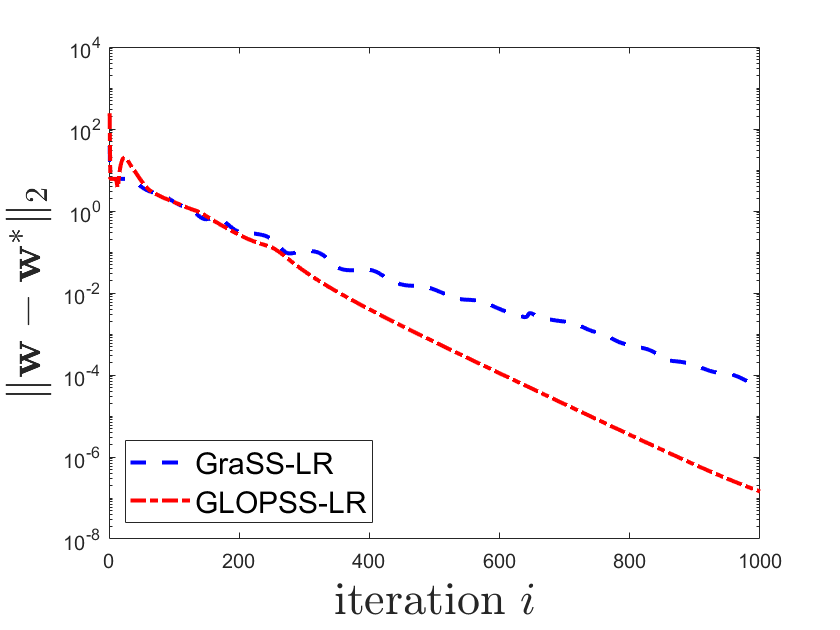}
		\caption{PA graph ($o=50$, $h=1$, $n=400$)}
		\label{P21}
	\end{subfigure}

    \begin{subfigure}{0.3\linewidth}
        \centering
        \includegraphics[width=0.9\linewidth]{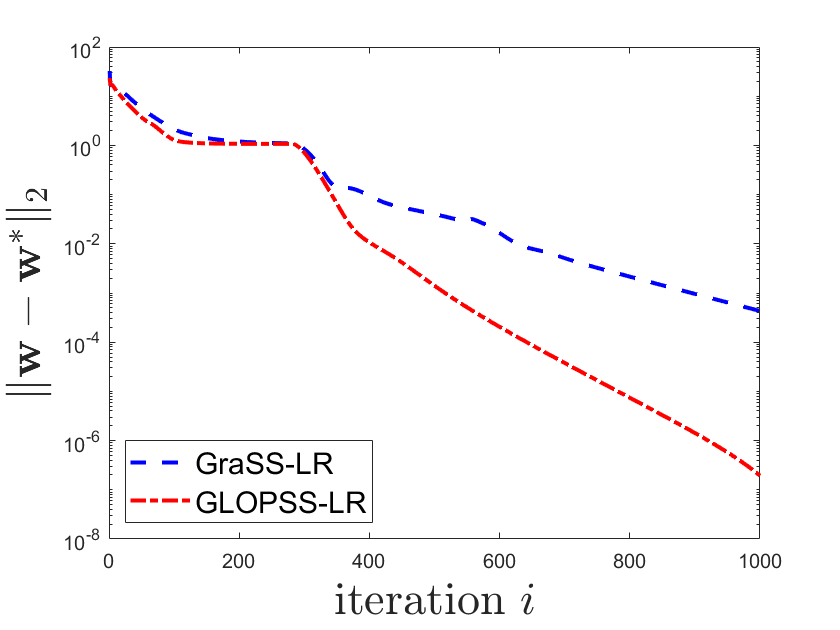}
        \caption{blumenau-drug graph ($o=70$, $h=5$, $n=100$)}
        \label{G22}
    \end{subfigure}
    \centering
    \begin{subfigure}{0.3\linewidth}
        \centering
        \includegraphics[width=0.9\linewidth]{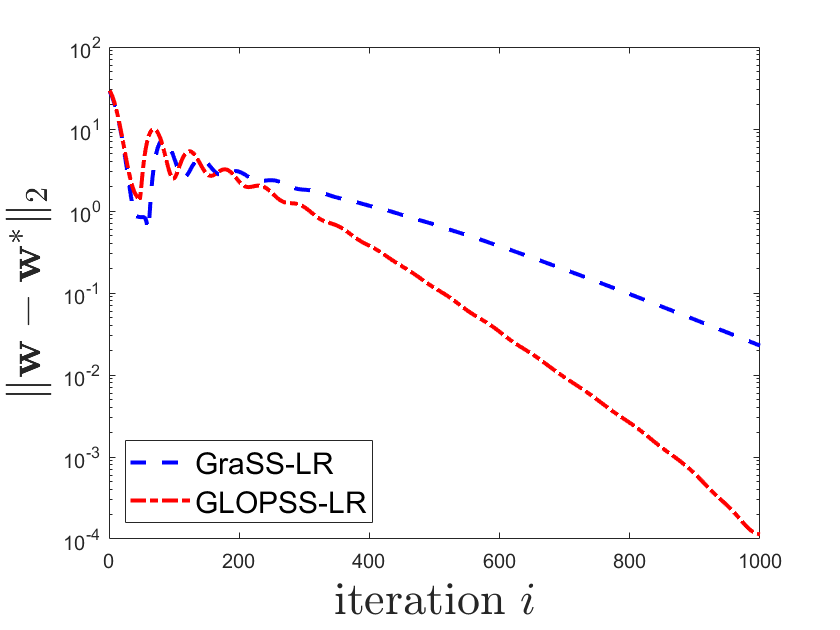}
        \caption{adjnoun graph ($o=107$, $h=5$, $n=100$)}
        \label{E22}
    \end{subfigure}
    \centering
    \begin{subfigure}{0.32\linewidth}
        \centering
        \includegraphics[width=0.9\linewidth]{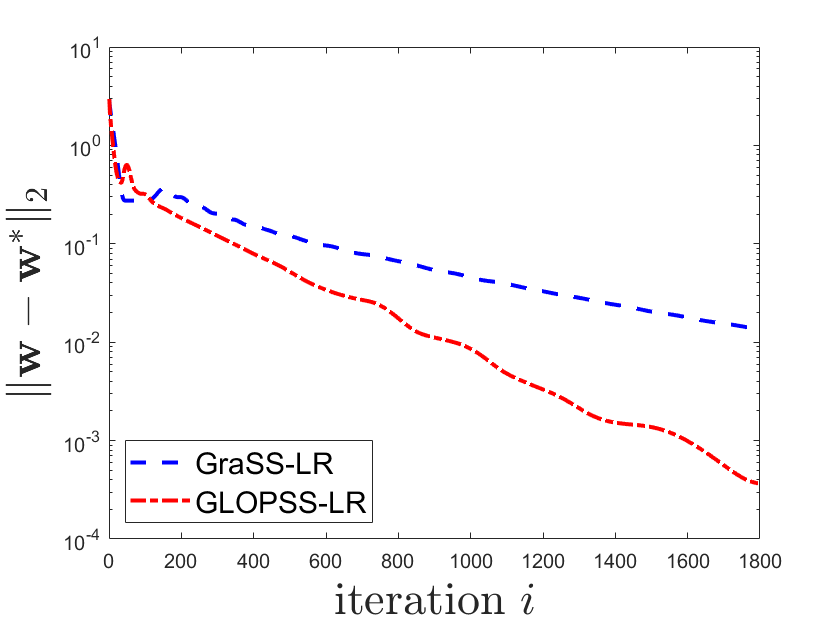}
        \caption{cintestinals graph ($o=200$, $h=5$, $n=1000$)}
        \label{P22}
    \end{subfigure}
	\caption{Convergence performance of GLOPSS-LR for graph learning on synthetic graphs}
	\label{fig1}
\end{figure*}

\subsection{Graph Learning Experiments}
\textit{1) Experimental Setup:} Following \cite{dong2016learning}, we generate smooth signals $\mathbf{X}$, using the eigen-decomposition of the graph Laplacian $\mathbf{L}=\mathbf{V}\Lambda\mathbf{V}^T$, where $\mathbf{L}=\text{Diag}(\mathbf{W1})-\mathbf{W}$. The signals matrix $\mathbf{X}=\mathbf{VZ}+\epsilon$ is composed of $\mathbf{Z}\in\mathbb{R}^{m\times n}$, drawn from a multivariate Gaussian distribution $\mathbf{Z}\sim\mathcal{N}(\mathbf{0},\Lambda^\dagger)$, where $\Lambda^\dagger$ is the pseudo-inverse of matrix $\Lambda$, and Gaussian noise $\epsilon\sim\mathcal{N}(\mathbf{0},\sigma\mathbf{I})$ with noise level $\sigma$. We evaluate recovery performance using weighted graphs and employ the $F_{score}$:
\begin{equation*}
    F_{\text{score}}=2\cdot\frac{percision\cdot recall}{percision+recall},
\end{equation*}
where $percision$ is the fraction of estimated edges that match the ground-truth edges and $recall$ the fraction of actual edges correctly identified.

\textit{2) Synthetic Graph:} We experiment with three types of synthetic graphs: the Gaussian (GA) graph, the Erd\"{o}s-R\'{e}nyi (ER) graph, the preferential attachment (PA) graph. In the Gaussian graph, nodes are randomly placed in a unit square, with edges added based on a radial basis function with a threshold of $0.75$. For example, an edge is placed between node $i$ and $j$ if the weight determined by the radial basis function $\exp(-d(i,j)^2/2\theta^2)$, where $d(i,j)$ is the Euclidean distance between nodes $i$ and $j$ and $\theta=0.5$ is the kernel width parameter, is at least $0.75$. The ER graph connects nodes independently with probability $0.2$, while the PA graph adds new nodes, connecting each to a previous node with probability proportional to previous node's degree. The edges in the Gaussian graph have weights given by the radial basis function, while those in the ER and PA are set to $1$. We use a noise level of $\sigma=0.5$ for all graphs.

Given the $\mathcal{O}(m^2)$ per-iteration computational cost of the algorithms, with $m$ being the dimension of graph signals, we measure their performance through the suboptimality gap $\Vert\mathbf{w}^i-\mathbf{w}^\ast\Vert_2$ for various $m$ and $n$ (number of graph signals). The average numerical performance of GLOPSS-LR for different graphs over 10 random testings is displayed and summarized in Fig. \ref{fig1}. As shown in Fig. \ref{fig1}, GLOPSS-LR converges noticeably more rapidly than GraSS-LR across all graph types. For instance, with the ER graph of $m=21$, GraSS-LR requires approximately $350$ iterations to achieve suboptimality gap $\Vert\mathbf{w}-\mathbf{w}^\ast\Vert$ below  $10^{-5}$, while GLOPSS-LR does so in about $150$ iterations. This aligns with our theoretical findings, as GLOPSS-LR not only allows for larger step sizes but also enables subsequent variables to exploit the most recent updates of preceding variables during each iteration, thus converging more rapidly than methods that consider variable partitioning.

\textit{3) Real-World Graphs:} We also test the numerical performance of the algorithms on several real-world graphs from Netzschleuder \footnote{\href{URL}{https://networks.skewed.de/}}: the Blumenau-drug network \cite{brattig2019city} with $m=75$, the Adjnoun network \cite{PhysRevE.74.036104} with $m=112$ and the Cintestinalis network \cite{ryan2016cns} with $m=205$. The numerical results, shown in Fig. \ref{fig1} indicate that GLOPSS-LR converges much faster than GraSS-LR in most cases. Even for the relatively larger Cintestinalis network, GLOPSS-LR can still achieve precision of $10^{-4}$, more quickly than GraSS-LR.
\begin{table*}[tbp]
    \caption{Runtime of Algorithms for Graph Learning from Smooth Signals Under Partial Observability}
    \label{tab.1}
    \centering
    \begin{subtable}[b]{1\linewidth}
        \centering
        \begin{tabular}{ccccccc}
            \toprule
            & o & Gsm-GL&Gsm-LR & GraSS-LR & GLOPSS-CS & GLOPSS-LR \\
            \midrule
            \multirow{2}{*}{Gaussian} & 20 & 2.1 &1.9 & 0.0023 & 0.0019 & \textbf{0.00185} \\
             & 50 & 15.5 & 14.7 & 0.1198 & \textbf{0.0803} & 0.0894  \\
            \midrule
            \multirow{2}{*}{ER} & 20 & 4.3 &4.8 &0.00828 & \textbf{0.0064}  & 0.007\\
             & 50 & 13.8 & 13.1 & 0.1058 &\textbf{0.0605} &0.0613 \\
            \midrule
            \multirow{2}{*}{PA} & 20 & 2.3 &2.6 & 0.0131& \textbf{0.0014} & 0.0046\\
             & 50 & 13.2 & 12.84& 0.1159 & \textbf{0.05} & 0.0519 \\
            \bottomrule
        \end{tabular}
        \label{tab1.a}
        \caption{Runtime (in seconds) on synthetic graphs}
    \end{subtable}
    
    \begin{subtable}[b]{1\linewidth}
        \centering
        \begin{tabular}{ccccccc}
            \toprule
            & o & Gsm-GL& Gsm-LR &GraSS-LR & GLOPSS-CS & GLOPSS-LR\\
            \midrule
            blumenau & 70 & 16.84 &15.9 & 0.248 &\textbf{0.083} & 0.098\\
            \midrule
            adjnoun & 107 & 474.8 & 443.6 & 1.043 & \textbf{0.665} & 0.72 \\
            \bottomrule
        \end{tabular}
        \label{tab1.b}
        \caption{Runtime (in seconds) on real-world graphs}
    \end{subtable}  
\end{table*}
\subsection{Runtime Comparison}
Next, we will show the runtime comparison of GLOPSS, GraSS and Gsm-GL (Gsm-LR) proposed by \cite{buciulea2022learning}, who use CVX as a solver. We evaluate the CPU runtime of GLOPSS-CS and GLOPSS-LR algorithms by terminating it when the suboptimality gap $\Vert\mathbf{w}^i-\mathbf{w}^\ast\Vert_2$ falls below $10^{-6}$. Given that Problem \eqref{eq11} is convex, we use the CVX package \cite{grant2014cvx} with SDPT3 as the solver and set the precision to the highest available level. The runtime of Gsm-GL and Gsm-LR are provided as baselines for comparison. 

Table \ref{tab.1}. presents the average runtime across $10$ independent trials under the same experiment settings as those used in Fig. \ref{fig1}. The results indicate that GLOPSS-LR and GLOPSS-CS generally require less runtime compared to GraSS-LR. Notably, for the medium-sized networks such as Adjnoun and Blumenau-drug networks, GLOPSS-LR and GLOPSS-CS show a significant reduction in runtime compared to both CVX and GraSS-LR. We do not provide runtime comparison for the largest Cintestinalis graph, as all the compared methods except GLOPSS-LR and GLOPSS-CS exhibit exceedingly slowly convergence. We can also find that GLOPSS-CS and GLOPSS-LR have the similar performance. The choice of two algorithms depends on the specific needs of the application, and there is no definitive answer as to which is superior or inferior.
\begin{figure}[htbp]
    \centering
    \includegraphics[width=0.5\linewidth]{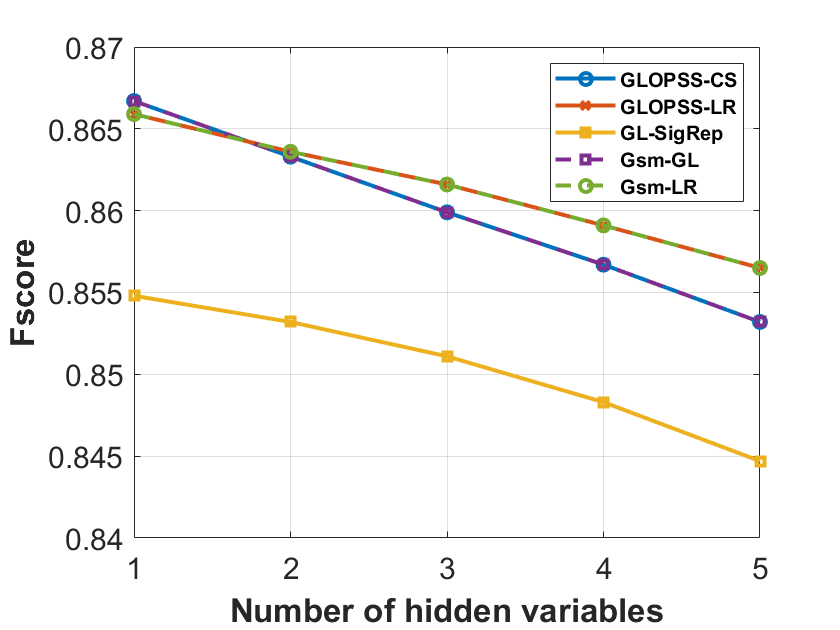}
    \caption{Median $F_{\text{score}}$ for GLOPSS-GL and GLOPSS-LR based on smooth siganls with $m=25$ and $n=100$ while the number of hidden variables $h$ varing from $1$ to $5$ when using Gaussian graphs.}
    \label{pict.5}
\end{figure} 
\subsection{Influence of Latent Nodes}
Fig. \ref{pict.5} illustrates the variation in $F_{\text{score}}$ as the number of hidden nodes $h$ increases, for various recovery algorithms. Graphs are randomly generated using the model in \cite{dong2016learning}, with hidden nodes chosen uniformly at random from the entire graph. The algorithms tested include: (i) GL-SigRep, as described in \cite{dong2016learning}; (ii) Gsm-GL and Gsm-LR, as described in \cite{buciulea2022learning}; (iii) GLOPSS-CS, as described in Algorithm \ref{alg1}, which employs column-sparsity promotion in $\tilde{\mathbf{K}}$ via group LASSO with $\gamma_\ast=0$; and (iv) GLOPSS-LR, the low-rank regularized algorithm from Algorithm \ref{alg2}. 

Comparing GL-SigRep with GLOPSS-LR allows us to assess the benefit of incorporating hidden variables into the model. Results from Fig. \ref{pict.5} show that, although all algorithms experience performance deteriorate as the number of hidden variables increases, GLOPSS-CS and GLOPSS-LR, which account for the presence of hidden variables, outperform the alternatives. Their performance decline is  slower with increasing $h$, demonstrating the importance of accounting for hidden variables. The observed overall performance decline is expected as more hidden variables renders the topology inference problem more challenging and ill-posed. Comparing GLOPSS-CS with GLOPSS-LR, their performances are similar, reflecting the sparsity of the generated graphs. We also found that the performance of Gsm-LR (resp. Gsm-GL) and GLOPSS-LR (resp. GLOPSS-CS) is almost identical, which is as expected since the optimal solutions of the two algorithms are the same. Our proposed GLOPSS algorithm, which converges linearly, simply offers faster computation, enabling quicker convergence to this single optimal solution.

\begin{figure}[htbp]
    \centering
    \includegraphics[width=0.5\linewidth]{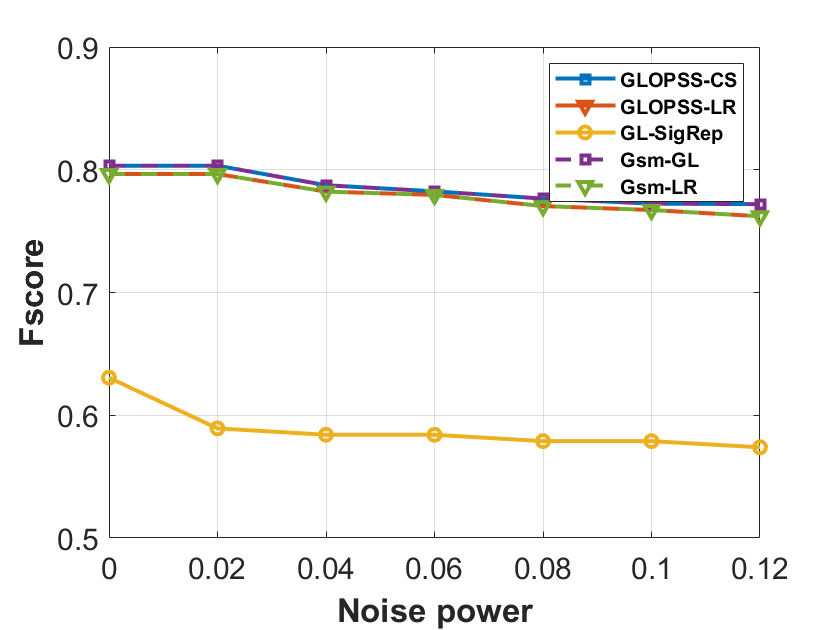}
    \caption{$F_{\text{score}}$ for GLOPSS-GL and GLOPSS-LR based on smooth siganls with $m=25$ and $n=100$ while the noise level present in the observations $\mathbf{X}$ when using ER graphs with link probability $p=0.2$.}
    \label{pict.6}
\end{figure} 
\subsection{Robustness}
The experiments assumes that the observations $\mathbf{X}_{\mathcal{O}}$ correspond to the ground-truth signals corrupted by additive white Gaussian noise. For experiments with additive white Gaussian noise, we evaluate the link-identification performance by measuring the $F_{\text{score}}$ of GL-SigRep, GLOPSS-CS and GLOPSS-LR as noise power increases. Using ER graphs with edge probability $p=0.2$ and one hidden variables ($h=1$), results in Fig. \ref{pict.6} reveal that the performance declines with increasing noise level. Despite not being specifically designed to handle noisy observations, the proposed algorithms exhibit a slower rate of $F_{\text{score}}$ degradation relative to the increase in noise power, demonstrating their robustness. Moreover, our algorithm consistently outperform GL-SigRep, highlighting the advantage of considering hidden nodes in  the recovery process.

\section{Conclusion}\label{sect6}
In this article, we present a highly efficient and versatile optimization approach for network topology inference from smooth signals under partial observability. We have demonstrated that our method achieves linear convergence to the optimal solution by showing that the KKT points of the formulation possess an error bound property. Extensive numerical experiments, conducted on both synthetic and real-world datasets, confirm that our algorithms outperform existing state-of-the-art methods in terms of both convergence speed and computational efficiency. 

Despite the strong theoretical foundations and empirical performance demonstrated by our first-order framework for network topology inference under partial observability, several promising directions remain open for further exploration: (1) Extending the proposed methods to handle time-varying networks would enable the inference of evolving topologies from streaming smooth signals. In such scenarios, algorithms must balance adaptivity and stability, perhaps through online optimization techniques or recursive updates that exploit temporal smoothness. (2) Our current analysis assumes smoothness of signals with respect to the underlying graph. Future work could relax this assumption to accommodate piecewise-smooth, sparse, or heavy-tailed signal distributions, potentially via composite regularizers or robust divergence measures that retain convergence guarantees. (3) Finally, tailoring and validating the framework for domain-specific applications—such as brain connectomics, power-grid monitoring, or financial network analysis—will test its versatility. Each domain may introduce unique constraints (e.g., anatomical priors in neuroscience or physical laws in engineering) that inspire novel algorithmic adaptations.

\section*{Acknowledgments}
This work was supported in part by Sichuan Youth Science and Technology Innovation Team (Grant Nos. 2022JDTD0014, 2021JDJQ0036). The work of Zhiguo Wang was supported in part by the National Natural Science Foundation of China under Grant 62203313.

\appendix
\section{Appendix}\label{sec9}
\renewcommand{\theequation}{A.\arabic{equation}} 
\setcounter{equation}{0} 

\subsection{Proof of \textit{Proposition 1}}
\label{app:prop1}
\textit{Proof}. Note that $f_{1}(\mathbf{w})=\frac{1}{2}\mathbf{z}^T\mathbf{w}+\beta\Vert\mathbf{w}\Vert_2^2+\mathbb{I}_{\mathbb{R}_+^p}(\mathbf{w})$, and
\begin{align*}
    \text{prox}_{\tau_1f_1}(\mathbf{w})&=\text{arg}\min_{\mathbf{x}}\left\{f_1(\mathbf{x})+\frac{1}{2\tau_1}\Vert\mathbf{x}-\mathbf{w}\Vert_2^2\right\}\\
    &=\text{arg}\min_{\mathbf{x}}\left\{\frac{1}{2}\mathbf{z}^T\mathbf{x}+\beta\Vert\mathbf{x}\Vert_2^2+\mathbb{I}_{\mathbb{R}_+^p}(\mathbf{x})+\frac{1}{2\tau_1}\Vert\mathbf{x}-\mathbf{w}\Vert_2^2\right\}\\
    &=\max\left\{\frac{\mathbf{w}-\frac{1}{2}\tau_1\mathbf{z}}{2\tau_1\beta+1}, \mathbf{0}\right\},
\end{align*}
where $\max\{\mathbf{a}, \mathbf{0}\}$ denotes the element-wise maximum of $\mathbf{a}$ and $\mathbf{0}$. This completes the proof. $\hfill\square$

\subsection{Proof of \textit{Proposition 2}}
\label{app:prop2}
\textit{Proof}. Note that $f_3(\mathbf{k})=2\mathbf{b}^T\mathbf{k}+\gamma_{2,1}h(\mathbf{k})$, where $h(\mathbf{k})=\Vert\tilde{\mathbf{K}}\Vert_{2,1}$, and combining Lemma 2, we have
\begin{align*}
    \text{prox}_{\tau_3f_3}(\mathbf{k})&=\text{arg}\min_{\mathbf{x}}\left\{f_3(\mathbf{x})+\frac{1}{2\tau_3}\Vert\mathbf{x}-\mathbf{k}\Vert_2^2\right\}\\
    &=\text{arg}\min_{\mathbf{x}}\left\{2\mathbf{b}^T\mathbf{x}+\gamma_{2,1}\Vert\tilde{\mathbf{X}}\Vert_{2,1}+\frac{1}{2\tau_3}\Vert\mathbf{x}-\mathbf{k}\Vert_2^2\right\}\\
    &=\left[1-\frac{\tau_3\gamma_{2,1}}{\Vert\mathbf{k}-2\tau_3\mathbf{b}\Vert_2}\right]_+(\mathbf{k}-2\tau_3\mathbf{b}).
\end{align*}
This completes the proof. $\hfill\square$

\subsection{Proof of \textit{Proposition 3}}
\label{app:prop3}
\textit{Proof}. Note that $f_4(\mathbf{v})=\mathbf{d}^T\mathbf{v}+\mathbb{I}_{\mathbb{R}_+}(\mathbf{v})$, thus, we have
\begin{align*}
    \text{prox}_{\tau_4f_4}(\mathbf{v})&=\text{arg}\min_{\mathbf{x}}\left\{f_4(\mathbf{x})+\frac{1}{2\tau_4}\Vert\mathbf{x}-\mathbf{v}\Vert_2^2\right\}\\
    &=\text{arg}\min_{\mathbf{x}}\left\{\mathbf{d}^T\mathbf{x}+\mathbb{I}_{\mathbb{R}_+}(\mathbf{x})+\frac{1}{2\tau_4}\Vert\mathbf{x}-\mathbf{v}\Vert_2^2\right\}\\
    &=\max\{\mathbf{v}-\tau_4\mathbf{d}, \mathbf{0}\}.
\end{align*}
This completes the proof. $\hfill\square$

\subsection{Proof of \textit{Theorem 1}}
\label{app:thm_1}
\textit{Proof}. Our Theorem \ref{thm_1} is a direct derivation of \cite{hong2017linear}. The main convergence result, Theorem 3.1 in \cite{hong2017linear}, is based on Assumption $(a)$-$(g)$ in \cite{hong2017linear}. So, to prove our Theorem \ref{thm_1}, we only need to verify whether our Problem \eqref{eq11} satisfies these assumptions. 

First, it is easy to verify that Problem \eqref{eq11} meets Assumption $(a)$, $(c)$, $(d)$ and $(e)$. And next, Assumption $(b)$ in \cite{hong2017linear} says that $f(\mathbf{x})=f_1(\mathbf{x}_1)+\ldots+f_K(\mathbf{x}_K)$, with each $f_k$ further decomposable as 
\begin{align}
    f_k(\mathbf{x}_k)=g_k(\mathbf{A}_k\mathbf{x}_k)+h(\mathbf{x}_k).
    \label{equa34}
\end{align}
Although our $f_i$, $i= 1, 2, 3, 4$ only consist of the convex function $f_i$, this does not contradict Equation \eqref{equa34}, for the strictly convex part being effectively absent by having $\mathbf{A}_k=\mathbf{0}$.

And then, for Assumption $(f)$, although our matrix $\mathbf{A}$ does not have full rank, like the discussion at Section 4.1 in \cite{hong2017linear}, our algorithms align with linearized proximal ADMM framework rather the classic ADMM framework. Thus, without requiring the full rankness of $\mathbf{A}$, we can still establish the linear convergence of our GLOPSS-CS and GLOPSS-LR.

Last but not least, let $\mathbf{X}=X_1\times X_2\times X_3\times X_4$ be the feasible set of Problem \eqref{eq11}, with $\mathbf{w}\in X_1$, $\mathbf{u}\in X_2$, $\mathbf{k}\in X_3$, and $\mathbf{v}\in X_4$. For Assumption $(g)$, it is easy to verify that the feasible sets $X_i$, $i=1,2,3,4$ are compact polyhedral sets, which are needed to ensure that certain error bound of the primal and dual problem of \eqref{eq11} hold. For example, for Algorithm \ref{alg1}, $f_3(\mathbf{k})$ includes the group LASSO penalization $\Vert\cdot\Vert_{2,1}$, the proof of error bound of this type of function can be found in \cite{zhang2013linear}. 

Thus, we have verified that Problem \eqref{eq11} meets all the conditions listed in Assumption $(a)$–$(g)$ in \cite{hong2017linear}, hence the linear convergence result can be obtained. This completes the proof of Theorem \ref{thm_1}. $\hfill\square$

\subsection{Proof of \textit{Theorem 2}}
\label{app:thm_2}
\textit{Proof}. Since $(\mathbf{w}^\ast,\mathbf{u}^\ast, \mathbf{k}^\ast, \mathbf{v}^\ast,\lambda^\ast)$ is optimal to \eqref{eq11}, it follows from the KKT conditions that the following hold:
\begin{align}
    \mathbf{0} &\in \partial f_1(\mathbf{w}^\ast)-\mathbf{M}_1^T\lambda^\ast, \label{equa1}\\
    \mathbf{0} &\in \partial f_2(\mathbf{u}^\ast)-\mathbf{M}_2^T\lambda^\ast, \label{equa2}\\
    \mathbf{0} &\in \partial f_3(\mathbf{k}^\ast)-\mathbf{M}_3^T\lambda^\ast, \label{equa3}\\
    \mathbf{0} &\in \partial f_4(\mathbf{v}^\ast)-\mathbf{M}_4^T\lambda^\ast, \label{equa4}\\
    \mathbf{0} &= \mathbf{Ax}. \label{equa5}
\end{align}

Note that the optimality  conditions for the first subproblem (i.e., the subproblem with respect to $\mathbf{w}$) in Algorithm \ref{alg1} is given by
\begin{align}
    \mathbf{0}&\in\frac{\tau_1}{\rho}\partial f_1(\mathbf{w}^{i+1})+\mathbf{w}^{i+1}-\mathbf{w}^{i}+\tau_1\mathbf{M}_1^T(\mathbf{M}_1\mathbf{w}^i+\mathbf{M}_2\mathbf{u}^i+\mathbf{M}_3\mathbf{k}^i+\mathbf{M}_4\mathbf{v}^i-\frac{\lambda^i}{\rho}).
    \label{equa6}
\end{align}
By using the updating formula for $\lambda^i$, i.e.,
\begin{equation}
    \lambda^{i+1}=\lambda^i-\rho\mathbf{Ax}^{i+1},
    \label{equa7}
\end{equation}
\eqref{equa6} can be reduced to
\begin{align}
    \mathbf{0} &\in \frac{\tau_1}{\rho}\partial f_1(\mathbf{w}^{i+1})+\mathbf{w}^{i+1}-\mathbf{w}^{i} +\tau_1\mathbf{M}_1^T(\mathbf{M}_1(\mathbf{w}^i-\mathbf{w}^{i+1})\nonumber\\
    &\quad+\mathbf{M}_2(\mathbf{u}^i-\mathbf{u}^{i+1})+\mathbf{M}_3(\mathbf{k}^i-\mathbf{k}^{i+1})+\mathbf{M}_4(\mathbf{v}^i-\mathbf{v}^{i+1})-\frac{\lambda^{i+1}{}}{\rho}).
    \label{equa8}
\end{align}
Combining \eqref{equa1} and \eqref{equa8} and using the fact that $\partial f_1(\cdot)$ is a monotone operator, we get
\begin{align}
    &(\mathbf{w}^{i+1}-\mathbf{w}^\ast)^T(\frac{\rho}{\tau_1}(\mathbf{w}^i-\mathbf{w}^{i+1})-\rho\mathbf{M}_1^T\mathbf{M}_1(\mathbf{w}^i-\mathbf{w}^{i+1}) \nonumber\\
    &\quad-\rho\mathbf{M}_1^T(\mathbf{M}_2(\mathbf{u}^i-\mathbf{u}^{i+1})+\mathbf{M}_3(\mathbf{k}^i-\mathbf{k}^{i+1}) \nonumber\\
    &\quad+\mathbf{M}_4(\mathbf{v}^i-\mathbf{v}^{i+1}))+\mathbf{M}_1^T(\lambda^{i+1}-\lambda^\ast))\geq\mathbf{0}.
    \label{equa9}
\end{align}

The optimality conditions for the second subproblem (i.e., the subproblem with respect to $\mathbf{u}$) in Algorithm \ref{alg1} are given by
\begin{align}
    \mathbf{0}&\in\frac{\tau_2}{\rho}\partial f_2(\mathbf{u}^{i+1})+\mathbf{u}^{i+1}-\mathbf{u}^{i}+\tau_2\mathbf{M}_2^T(\mathbf{M}_1\mathbf{w}^{i+1}+\mathbf{M}_2\mathbf{u}^i+\mathbf{M}_3\mathbf{k}^i+\mathbf{M}_4\mathbf{v}^i-\frac{\lambda^i}{\rho}).
    \label{equa10}
\end{align}
Using \eqref{equa7}, \eqref{equa10} can be reduced to
\begin{align}
    &\mathbf{0}\in\frac{\tau_2}{\rho}\partial f_2(\mathbf{u}^{i+1})+\mathbf{u}^{i+1}-\mathbf{u}^i+\tau_2\mathbf{M}_2^T(\mathbf{M}_1\mathbf{w}^{i+1}+\mathbf{M}_2\mathbf{u}^i+\mathbf{M}_3\mathbf{k}^i+\mathbf{M}_4\mathbf{v}^i-\frac{\lambda^i}{\rho}).
    \label{equa11}
\end{align}
Combining \eqref{equa2} and \eqref{eq11} and using the fact that $\partial f_2(\cdot)$ is a monotone operator, we get 
\begin{align}
    &(\mathbf{u}^{i+1}-\mathbf{u}^\ast)^T(\frac{\rho}{\tau_2}(\mathbf{u}^i-\mathbf{u}^{i+1})-\rho\mathbf{M}_2^T\mathbf{M}_2(\mathbf{u}^i-\mathbf{u}^{i+1})-\mathbf{M}_2^T(\mathbf{M}_3(\mathbf{k}^i-\mathbf{k}^{i+1})+\mathbf{M}_4(\mathbf{v}^i-\mathbf{v}^{i+1}))\nonumber\\
    &\quad+\mathbf{M}_2^T(\lambda^{i+1}-\lambda^\ast))\geq\mathbf{0}.
    \label{equa12}
\end{align}
Similarly, for $\mathbf{k}$ and $\mathbf{v}$, we respectively have
\begin{align}
    (\mathbf{k}^{i+1}-\mathbf{k}^\ast)^T(\frac{\rho}{\tau_3}(\mathbf{k}^i-\mathbf{k}^{i+1})-\rho\mathbf{M}_3^T\mathbf{M}_3(\mathbf{k}^i-\mathbf{k}^{i+1})-\mathbf{M}_3^T(\mathbf{M}_4(\mathbf{v}^i-\mathbf{v}^{i+1}))+\mathbf{M}_2^T(\lambda^{i+1}-\lambda^\ast))\geq\mathbf{0}
    \label{equa13}
\end{align}
and
\begin{align}
    (\mathbf{v}^{i+1}-\mathbf{v}^\ast)^T(\frac{\rho}{\tau_4}(\mathbf{v}^i-\mathbf{v}^{i+1})-\rho\mathbf{M}_4^T\mathbf{M}_4(\mathbf{v}^i-\mathbf{v}^{i+1})+\mathbf{M}_4^T(\lambda^{i+1}-\lambda^\ast))\geq\mathbf{0}.
    \label{equa14}
\end{align}

Summing \eqref{equa9}, \eqref{equa12}, \eqref{equa13} and \eqref{equa14}, by using $\mathbf{Ax}^\ast=\mathbf{0}$, we obtain
\begin{align}
    &(\mathbf{w}^{i+1}-\mathbf{w}^{\ast})^T(\frac{\rho}{\tau_1}\mathbf{I}-\rho\mathbf{M}_1^T\mathbf{M}_1)(\mathbf{w}^i-\mathbf{w}^{i+1})+(\mathbf{u}^{i+1}-\mathbf{u}^{\ast})^T(\frac{\rho}{\tau_2}\mathbf{I}-\rho\mathbf{M}_2^T\mathbf{M}_2)(\mathbf{u}^i-\mathbf{u}^{i+1})\nonumber\\
    &\quad+(\mathbf{k}^{i+1}-\mathbf{k}^{\ast})^T(\frac{\rho}{\tau_3}\mathbf{I}-\rho\mathbf{M}_3^T\mathbf{M}_3)(\mathbf{k}^{i}-\mathbf{k}^{i+1})+\frac{\rho}{\tau_4}(\mathbf{v}^{i+1}-\mathbf{v}^\ast)^T(\mathbf{v}^i-\mathbf{v}^{i+1})\nonumber\\
    &\quad-(\lambda^i-\lambda^{i+1})^T\mathbf{M}_4(\mathbf{v}^i-\mathbf{v}^{i+1})+\frac{1}{\rho}(\lambda^i-\lambda^{i+1})^T(\lambda^{i+1}-\lambda^\ast)\geq \mathbf{0}.
    \label{equa15}
\end{align}
Using the notation of $\mathbf{y}^i$, $\mathbf{y}^\ast$ and $\mathbf{M}=\text{Diag}[\frac{\rho}{\tau_1}\mathbf{I}-\rho\mathbf{M}_1^T\mathbf{M}_1;\frac{\rho}{\tau_2}\mathbf{I}-\rho\mathbf{M}_2^T\mathbf{M}_2;\frac{\rho}{\tau_3}\mathbf{I}-\rho\mathbf{M}_3^T\mathbf{M}_3;\frac{\rho} {\tau_4}\mathbf{I};\frac{1}{\rho}\mathbf{I}]$, \eqref{equa15} can be written as
\begin{align}
    \langle \mathbf{y}^{i+1}-\mathbf{y}^\ast,\mathbf{y}^i-\mathbf{y}^{i+1}\rangle_{\mathbf{M}}\geq\langle \lambda^i-\lambda^{i+1}, \mathbf{M}_4\mathbf{v}^i-\mathbf{M}_4\mathbf{v}^{i+1}\rangle,
    \label{equa16}
\end{align}
which can be further written as 
\begin{align}
    \langle \mathbf{y}^{i}-\mathbf{y}^\ast,\mathbf{y}^i-\mathbf{y}^{i+1}\rangle_{\mathbf{M}}\geq\Vert\mathbf{y}^i-\mathbf{y}^{i+1}\Vert_{\mathbf{M}}^2+\langle \lambda^i-\lambda^{i+1}, \mathbf{M}_4\mathbf{v}^i-\mathbf{M}_4\mathbf{v}^{i+1}\rangle.
    \label{equa17}
\end{align}
Combining \eqref{equa17} with the identity
\begin{align}
    \Vert\mathbf{y}^{i+1}-\mathbf{y}^\ast\Vert_{\mathbf{M}}^2=\Vert\mathbf{y}^{i+1}-\mathbf{y}^i\Vert_{\mathbf{M}}^2-2\langle\mathbf{y}^i-\mathbf{y}^{i+1},\mathbf{y}^i-\mathbf{y}^\ast\rangle_{\mathbf{M}}+\Vert\mathbf{y}^i-\mathbf{y}^\ast\Vert_{\mathbf{M}}^2,
    \label{equa18}
\end{align}
we get 
\begin{align}
    \Vert\mathbf{y}^i-\mathbf{y}^\ast\Vert_{\mathbf{M}}^2-\Vert\mathbf{y}^{i+1}-\mathbf{y}^\ast\Vert_{\mathbf{M}}^2&=2\langle\mathbf{y}^i-\mathbf{y}^{i+1},\mathbf{y}^i-\mathbf{y}^\ast\rangle_{\mathbf{M}}-\Vert\mathbf{y}^{i+1}-\mathbf{y}^i\Vert_{\mathbf{M}}^2\nonumber\\
    &\geq\Vert\mathbf{y}^i-\mathbf{y}^{i+1}\Vert_{\mathbf{M}}^2+2\langle \lambda^i-\lambda^{i+1}, \mathbf{M}_4\mathbf{v}^i-\mathbf{M}_4\mathbf{v}^{i+1}\rangle.
    \label{equa19}
\end{align}

Let $\xi:=\frac{1}{2}+\frac{\tau_4\sigma_{\max}^2(\mathbf{M}_4)}{2}$, then we know that $\tau_4\sigma_{\max}^2(\mathbf{M}_4)<\xi<1$ since $\tau_4<\frac{1}{\sigma_{\max}^2(\mathbf{M}_4)}$. Let $\mu:=\frac{\xi}{\rho}$. Then from the Cauchy-Schwartz inequality we have
\begin{align}
    2\langle\lambda^i-\lambda^{i+1},\mathbf{M}_4\mathbf{v}^i-\mathbf{M}_4\mathbf{v}^{i+1}\rangle&\geq -\mu\Vert\lambda^i-\lambda^{i+1}\Vert^2-\frac{1}{\mu}\Vert\mathbf{M}_4\mathbf{v}^i-\mathbf{M}_4\mathbf{v}^{i+1}\Vert^2\nonumber\\
    &\geq -\mu\Vert\lambda^i-\lambda^{i+1}\Vert^2-\frac{1}{\mu}\sigma_{\max}^2(\mathbf{M}_4)\Vert\mathbf{M}_4\mathbf{v}^i-\mathbf{M}_4\mathbf{v}^{i+1}\Vert^2.
    \label{equa20}
\end{align}
Combining \eqref{equa19} and \eqref{equa20} we get
\begin{align}
    &\Vert\mathbf{y}^i-\mathbf{y}^\ast\Vert_{\mathbf{M}}^2-\Vert\mathbf{y}^{i+1}-\mathbf{y}^\ast\Vert_{\mathbf{M}}^2\nonumber\\
    &\quad\geq(\mathbf{w}^i-\mathbf{w}^{i+1})^T(\frac{\rho}{\tau_1}\mathbf{I}-\rho\mathbf{M}_1^T\mathbf{M}_1)(\mathbf{w}^i-\mathbf{w}^{i+1})+(\mathbf{u}^i-\mathbf{u}^{i+1})^T(\frac{\rho}{\tau_2}\mathbf{I}-\rho\mathbf{M}_2^T\mathbf{M}_2)(\mathbf{u}^i-\mathbf{u}^{i+1})\nonumber\\
    &\qquad+(\mathbf{k}^i-\mathbf{k}^{i+1})^T(\frac{\rho}{\tau_3}\mathbf{I}-\rho\mathbf{M}_3^T\mathbf{M}_3)(\mathbf{k}^i-\mathbf{k}^{i+1})+(\frac{\rho}{\tau_4}-\frac{1}{\mu}\sigma_{\max}^2(\mathbf{M}_4))\Vert\mathbf{v}^i-\mathbf{v}^{i+1}\Vert^2\nonumber\\
    &\qquad+(\frac{1}{\rho}-\mu)\Vert\lambda^i-\lambda^{i+1}\Vert^2\geq c\Vert\mathbf{y}^i-\mathbf{y}^{i+1}\Vert_{\mathbf{M}}^2,
    \label{equa21}
\end{align}
where $c:=\min\{\frac{\rho}{\tau_1}-\rho\sigma_{\max}^2(\mathbf{M}_1), \frac{\rho}{\tau_2}-\frac{1}{\mu}\sigma_{\max}^2(\mathbf{M}_2), \frac{\rho}{\tau_3}-\rho\sigma_{\max}^2(\mathbf{M}_3), \frac{\rho}{\tau_4}-\rho\sigma_{\max}^2(\mathbf{M}_4), \frac{1}{\rho}-\mu\}>0$. This completes the proof. $\hfill\square$

\subsection{Proof of \textit{Proposition 4}}
\label{app:prop4}
\textit{Proof}. We have $\mathbf{M}_4=\begin{bmatrix}
    \mathbf{a}^T \\
    \mathbf{0} 
\end{bmatrix}$. It is easy to have $\Vert\mathbf{M}_4\Vert_2=2$. 

Next, we have $\mathbf{M}_3=[2\mathbf{b}^T;\mathbf{0}]$, so we get
\begin{align*}
    \mathbf{M}_3\mathbf{M}_3^T=\begin{bmatrix}
        4\mathbf{b}^T\mathbf{b} & \mathbf{0}\\
        \mathbf{0} & \mathbf{0}
    \end{bmatrix},
\end{align*}
It is easy to have $\Vert\mathbf{M}_3\Vert_2=2\sqrt{o}$.
And then, we know $\mathbf{M}_2=[\mathbf{0};-\mathbf{I}]$, so we get $\Vert\mathbf{M}_2\Vert_2=1$.
Lastly, we have $\mathbf{M}_1=\begin{bmatrix}
    \frac{1}{2}\mathbf{z}^T \\
    \mathbf{B}
\end{bmatrix}$.  To bound $\Vert\mathbf{M}_1\Vert_2$, we have
\begin{align}
    \Vert\mathbf{M}_1\Vert_2\leq\Vert\frac{1}{2}\mathbf{z}^T\Vert_2+\Vert\mathbf{B}\Vert_2
    \label{equa31}
\end{align}

It is shown in \cite{saboksayr2021online} that $\Vert\mathbf{B}\Vert_2=\sqrt{2(o-1)}$. By the Gershgorin circle theorem, denoting $\kappa:=\max_{ij} z_{ij}$, we have
\begin{align}
    \lambda(\mathbf{zz}^T) \leq \kappa^2 o^2,\nonumber\\
    \label{equa32}
\end{align}
Combining \eqref{equa31} and \eqref{equa32}, gives
\begin{align}
    \Vert\mathbf{M}_1\Vert_2\leq\frac{1}{2}\kappa o+\sqrt{2(o-1)}
    \label{equa33}
\end{align}
This completes the proof. $\hfill\square$

\subsection{Proof of \textit{Lemma 5}}
\label{app:lemma5}
\textit{Proof}. By \eqref{eq15}, there exists $\bar{\mathbf{w}}\in\partial f_1(\mathbf{w}^{i+1})$ such that
\begin{align}
    \mathbf{w}^{i+1}&=P_{X_1}(\mathbf{w}^{i+1}-[\bar{\mathbf{w}}+\frac{\rho}{\tau_1}(\mathbf{w}^{i+1}-(\mathbf{w}^i-\tau_1\mathbf{M}_1^T(\mathbf{M}_1\mathbf{w}^i+\mathbf{M}_2\mathbf{u}^i+\mathbf{M}_3\mathbf{k}^i+\mathbf{M}_4\mathbf{v}^i-\frac{\lambda^i}{\rho})))]),
    \label{equ34}
\end{align}

Using the above relation, we can obtain that
\begin{align}
    &\text{dist}(\mathbf{0},\mathcal{E}_{\mathbf{w}}(\mathbf{y}^{i+1}))\nonumber\\ 
    &= \text{dist}(\mathbf{w}^{i+1},P_{X_1}(\mathbf{w}^{i+1}-(\partial f_1(\mathbf{w}^{i+1})-\mathbf{M}_1^T\lambda^{i+1}))) \nonumber\\
    &\leq \Vert P_{X_1}(\mathbf{w}^{i+1}-[\bar{\mathbf{w}}+\frac{\rho}{\tau_1}(\mathbf{w}^{i+1}-(\mathbf{w}^i-\tau_1\mathbf{M}_1^T(\mathbf{M}_1\mathbf{w}^i +\mathbf{M}_2\mathbf{u}^i+\mathbf{M}_3\mathbf{k}^i+\mathbf{M}_4\mathbf{v}^i-\frac{\lambda^i}{\rho})))])\nonumber\\
    &\quad-P_{X_1}(\mathbf{w}^{i+1}-(\bar{\mathbf{w}}-\mathbf{M}_1^T\lambda^{i+1}))\Vert \nonumber\\
    &\leq \Vert\frac{\rho}{\tau_1}(\mathbf{w}^{i+1}-\mathbf{w}^i)+\rho\mathbf{M}_1^T(\mathbf{M}_1\mathbf{w}^i+\mathbf{M}_2\mathbf{u}^i+\mathbf{M}_3\mathbf{k}^i+\mathbf{M}_4\mathbf{v}^i)+\mathbf{M}_1^T(\lambda^{i+1}-\lambda^i)\Vert \nonumber\\
    &=\Vert\frac{\rho}{\tau_1}(\mathbf{w}^{i+1}-\mathbf{w}^i)+\rho\mathbf{M}_1^T[\mathbf{M}_1(\mathbf{w}^{i}-\mathbf{w}^{i+1})+\mathbf{M}_2(\mathbf{u}^i-\mathbf{u}^{i+1})+\mathbf{M}_3(\mathbf{k}^i-\mathbf{k}^{i+1})\nonumber\\
    &\quad+\mathbf{M}_4(\mathbf{v}^i-\mathbf{v}^{i+1})]\Vert \nonumber\\
    &\leq \Vert\frac{\rho}{\tau_1}\mathbf{I}-\rho\mathbf{M}_1^T\mathbf{M}_1\Vert\Vert\mathbf{w}^{i+1}-\mathbf{w}^i\Vert+\rho\Vert\mathbf{M}_1^T\mathbf{M}_2\Vert\Vert\mathbf{u}^{i+1}-\mathbf{u}^{i}\Vert+\rho\Vert\mathbf{M}_1^T\mathbf{M}_3\Vert\Vert\mathbf{k}^{i+1}-\mathbf{k}^{i}\Vert\nonumber\\
    &\quad+\rho\Vert\mathbf{M}_1^T\mathbf{M}_4\Vert\Vert\mathbf{v}^{i+1}-\mathbf{v}^{i}\Vert,
    \label{equa35}
\end{align}
where the second equality holds due to the update formula of $\lambda^i$ in \eqref{eq23}. 

By \eqref{eq17}, there exists $\bar{\mathbf{u}}\in\partial f_2(\mathbf{u}^{i+1})$ such that
\begin{align}
    \mathbf{u}^{i+1}&=P_{X_2}(\mathbf{u}^{i+1}-[\bar{\mathbf{u}}+\frac{\rho}{\tau_2}(\mathbf{u}^{i+1}-(\mathbf{u}^i-\tau_2\mathbf{M}_1^T(\mathbf{M}_1\mathbf{w}^{i+1} +\mathbf{M}_2\mathbf{u}^i+\mathbf{M}_3\mathbf{k}^i+\mathbf{M}_4\mathbf{v}^i-\frac{\lambda^i}{\rho})))]).
    \label{equ36}
\end{align}
Then, like \eqref{equa35},  we have
\begin{align}
    &\text{dist}(\mathbf{0},\mathcal{E}_{\mathbf{u}}(\mathbf{y}^{i+1}))\nonumber\\
    &= \text{dist}(\mathbf{u}^{i+1},P_{X_2}(\mathbf{u}^{i+1}-(\partial f_2(\mathbf{u}^{i+1})-\mathbf{M}_2^T\lambda^{i+1}))) \nonumber\\
    &\leq \Vert P_{X_2}(\mathbf{u}^{i+1}-[\bar{\mathbf{u}}+\frac{\rho}{\tau_2}(\mathbf{u}^{i+1}-(\mathbf{u}^i-\tau_2\mathbf{M}_2^T(\mathbf{M}_1\mathbf{w}^{i+1}+\mathbf{M}_2\mathbf{u}^i+\mathbf{M}_3\mathbf{k}^i+\mathbf{M}_4\mathbf{v}^i-\frac{\lambda^i}{\rho})))])\nonumber\\
    &\quad-P_{X_2}(\mathbf{u}^{i+1}-(\bar{\mathbf{u}}-\mathbf{M}_2^T\lambda^{i+1}))\Vert \nonumber\\
    &\leq \Vert\frac{\rho}{\tau_2}(\mathbf{u}^{i+1}-\mathbf{u}^i)+\rho\mathbf{M}_2^T(\mathbf{M}_1\mathbf{w}^{i+1}+\mathbf{M}_2\mathbf{u}^i+\mathbf{M}_3\mathbf{k}^i+\mathbf{M}_4\mathbf{v}^i)+\mathbf{M}_2^T(\lambda^{i+1}-\lambda^i)\Vert \nonumber\\
    &=\Vert\frac{\rho}{\tau_2}(\mathbf{u}^{i+1}-\mathbf{u}^i)+\rho\mathbf{M}_2^T[\mathbf{M}_2(\mathbf{u}^{i}-\mathbf{u}^{i+1})+\mathbf{M}_3(\mathbf{k}^i-\mathbf{k}^{i+1})+\mathbf{M}_4(\mathbf{v}^i-\mathbf{v}^{i+1})]\Vert \nonumber\\
    &\leq \Vert\frac{\rho}{\tau_2}\mathbf{I}-\rho\mathbf{M}_2^T\mathbf{M}_2\Vert\Vert\mathbf{u}^{i+1}-\mathbf{u}^i\Vert+\rho\Vert\mathbf{M}_2^T\mathbf{M}_3\Vert\Vert\mathbf{k}^{i+1}-\mathbf{k}^{i}\Vert+\rho\Vert\mathbf{M}_2^T\mathbf{M}_4\Vert\Vert\mathbf{v}^{i+1}-\mathbf{v}^{i}\Vert,
    \label{equa37}
\end{align}

By \eqref{eq19}, there exists $\bar{\mathbf{k}}\in\partial f_3(\mathbf{k}^{i+1})$ such that
\begin{align}
    \mathbf{k}^{i+1}&=P_{X_3}(\mathbf{k}^{i+1}-[\bar{\mathbf{k}}+\frac{\rho}{\tau_3}(\mathbf{k}^{i+1}-(\mathbf{k}^i-\tau_3\mathbf{M}_3^T(\mathbf{M}_1\mathbf{w}^{i+1} +\mathbf{M}_2\mathbf{u}^{i+1}+\mathbf{M}_3\mathbf{k}^i+\mathbf{M}_4\mathbf{v}^i-\frac{\lambda^i}{\rho})))]).
    \label{equa38}
\end{align}
Then, similarly, we have
\begin{align}
    &\text{dist}(\mathbf{0},\mathcal{E}_{\mathbf{k}}(\mathbf{y}^{i+1}))\nonumber\\ 
    &= \text{dist}(\mathbf{k}^{i+1},P_{X_3}(\mathbf{k}^{i+1}-(\partial f_3(\mathbf{k}^{i+1})-\mathbf{M}_3^T\lambda^{i+1}))) \nonumber\\
    &\leq \Vert P_{X_3}(\mathbf{k}^{i+1}-[\bar{\mathbf{k}}+\frac{\rho}{\tau_3}(\mathbf{k}^{i+1}-(\mathbf{k}^i-\tau_3\mathbf{M}_3^T(\mathbf{M}_1\mathbf{w}^{i+1}+\mathbf{M}_2\mathbf{u}^{i+1}+\mathbf{M}_3\mathbf{k}^i+\mathbf{M}_4\mathbf{v}^i-\frac{\lambda^i}{\rho})))])\nonumber\\
    &\quad-P_{X_3}(\mathbf{k}^{i+1}-(\bar{\mathbf{k}}-\mathbf{M}_3^T\lambda^{i+1}))\Vert \nonumber\\
    &\leq \Vert\frac{\rho}{\tau_3}(\mathbf{k}^{i+1}-\mathbf{k}^i)+\rho\mathbf{M}_3^T(\mathbf{M}_1\mathbf{w}^{i+1}+\mathbf{M}_2\mathbf{u}^{i+1}+\mathbf{M}_3\mathbf{k}^i+\mathbf{M}_4\mathbf{v}^i)+\mathbf{M}_3^T(\lambda^{i+1}-\lambda^i)\Vert \nonumber\\
    &=\Vert\frac{\rho}{\tau_3}(\mathbf{k}^{i+1}-\mathbf{k}^i)+\rho\mathbf{M}_3^T[\mathbf{M}_3(\mathbf{k}^{i}-\mathbf{k}^{i+1})+\mathbf{M}_4(\mathbf{v}^i-\mathbf{v}^{i+1})]\Vert \nonumber\\
    &\leq \Vert\frac{\rho}{\tau_3}\mathbf{I}-\rho\mathbf{M}_3^T\mathbf{M}_3\Vert\Vert\mathbf{k}^{i+1}-\mathbf{k}^i\Vert+\rho\Vert\mathbf{M}_3^T\mathbf{M}_4\Vert\Vert\mathbf{v}^{i+1}-\mathbf{v}^{i}\Vert.
    \label{equa39}
\end{align}

By \eqref{eq21}, there exists $\bar{\mathbf{v}}\in\partial f_4(\mathbf{v}^{i+1})$ such that
\begin{align}
    \mathbf{v}^{i+1}=P_{X_4}(\mathbf{v}^{i+1}-[\bar{\mathbf{v}}+\frac{\rho}{\tau_4}(\mathbf{v}^{i+1}-(\mathbf{v}^i-\tau_4\mathbf{M}_4^T(\mathbf{M}_1\mathbf{w}^{i+1} +\mathbf{M}_2\mathbf{u}^{i+1}+\mathbf{M}_3\mathbf{k}^{i+1}+\mathbf{M}_4\mathbf{v}^i-\frac{\lambda^i}{\rho})))]).
    \label{equa40}
\end{align}
Then, like \eqref{equa35}, we have
\begin{align}
    &\text{dist}(\mathbf{0},\mathcal{E}_{\mathbf{v}}(\mathbf{y}^{i+1}))\nonumber\\
    &= \text{dist}(\mathbf{v}^{i+1},P_{X_4}(\mathbf{v}^{i+1}-(\partial f_4(\mathbf{v}^{i+1})-\mathbf{M}_4^T\lambda^{i+1}))) \nonumber\\
    &\leq \Vert P_{X_4}(\mathbf{v}^{i+1}-[\bar{\mathbf{v}}+\frac{\rho}{\tau_4}(\mathbf{v}^{i+1}-(\mathbf{v}^i-\tau_4\mathbf{M}_4^T(\mathbf{M}_1\mathbf{w}^{i+1} +\mathbf{M}_2\mathbf{u}^{i+1}+\mathbf{M}_3\mathbf{k}^{i+1}+\mathbf{M}_4\mathbf{v}^i\nonumber\\
    &\quad-\frac{\lambda^i}{\rho})))])-P_{X_4}(\mathbf{v}^{i+1}-(\bar{\mathbf{v}}-\mathbf{M}_4^T\lambda^{i+1}))\Vert \nonumber\\
    &\leq \Vert\frac{\rho}{\tau_4}(\mathbf{v}^{i+1}-\mathbf{v}^i)+\rho\mathbf{M}_4^T(\mathbf{M}_1\mathbf{w}^{i+1}+\mathbf{M}_2\mathbf{u}^{i+1}+\mathbf{M}_3\mathbf{k}^{i+1}+\mathbf{M}_4\mathbf{v}^i)+\mathbf{M}_4^T(\lambda^{i+1}-\lambda^i)\Vert \nonumber\\
    &=\Vert\frac{\rho}{\tau_4}(\mathbf{v}^{i+1}-\mathbf{v}^i)+\rho\mathbf{M}_4^T\mathbf{M}_4(\mathbf{v}^i-\mathbf{v}^{i+1})\Vert \nonumber\\
    &\leq \Vert\frac{\rho}{\tau_4}\mathbf{I}-\rho\mathbf{M}_4^T\mathbf{M}_4\Vert\Vert\mathbf{v}^{i+1}-\mathbf{v}^i\Vert.
    \label{equa41}
\end{align}
It is obvious that 
\begin{align}
    \Vert\mathcal{E}_\lambda(\mathbf{y}^{i+1})\Vert&=\Vert\mathbf{M}_1\mathbf{w}^{i+1}+\mathbf{M}_2\mathbf{u}^{i+1}+\mathbf{M}_3\mathbf{k}^{i+1}+\mathbf{M}_4\mathbf{v}^{i+1}\Vert\nonumber\\
    &=\frac{1}{\rho}\Vert\lambda^{i+1}-\lambda^i\Vert.
    \label{equa42}
\end{align}
By \eqref{equa35}, \eqref{equa37}, \eqref{equa39}, \eqref{equa41} and \eqref{equa42}, we have
\begin{align}
    &\text{dist}^2(\mathbf{0},\mathcal{E}(\mathbf{y}^{i+1}))\nonumber\\
    &= \text{dist}^2(\mathbf{0},\mathcal{E}_\mathbf{w}(\mathbf{y}^{i+1}))+ \text{dist}^2(\mathbf{0},\mathcal{E}_\mathbf{u}(\mathbf{y}^{i+1}))+ \text{dist}^2(\mathbf{0},\mathcal{E}_\mathbf{k}(\mathbf{y}^{i+1}))+ \text{dist}^2(\mathbf{0},\mathcal{E}_\mathbf{v}(\mathbf{y}^{i+1}))\nonumber\\
    &\quad+ \text{dist}^2(\mathbf{0},\mathcal{E}_\lambda(\mathbf{y}^{i+1}))\nonumber\\
    &\leq2\Vert\frac{\rho}{\tau_1}\mathbf{I}-\rho\mathbf{M}_1^T\mathbf{M}_1\Vert^2\Vert\mathbf{w}^{i+1}-\mathbf{w}^i\Vert^2+(2\rho^2\Vert\mathbf{M}_1^T\mathbf{M}_2\Vert^2+\Vert\frac{\rho}{\tau_2}\mathbf{I}-\rho\mathbf{M}_2^T\mathbf{M}_2\Vert^2)\Vert\mathbf{u}^{i+1}-\mathbf{u}^i\Vert^2\nonumber\\
    &\quad+(2\rho^2\Vert\mathbf{M}_1^T\mathbf{M}_3\Vert^2+2\rho^2\Vert\mathbf{M}_2^T\mathbf{M}_3\Vert^2+\Vert\frac{\rho}{\tau_3}\mathbf{I}-\rho\mathbf{M}_3^T\mathbf{M}_3\Vert^2)\Vert\mathbf{k}^{i+1}-\mathbf{k}^i\Vert^2\nonumber\\
    &\quad+(2\rho^2\Vert\mathbf{M}_1^T\mathbf{M}_4\Vert^2+2\rho^2\Vert\mathbf{M}_2^T\mathbf{M}_4\Vert^2+2\Vert\mathbf{M}_3^T\mathbf{M}_4\Vert^2+\Vert\frac{\rho}{\tau_4}\mathbf{I}-\rho\mathbf{M}_4^T\mathbf{M}_4\Vert^2)\Vert\mathbf{v}^{i+1}-\mathbf{v}^i\Vert^2\nonumber\\
    &\quad+\frac{1}{\rho^2}\Vert\lambda^{i+1}-\lambda^i\Vert^2.
\end{align}
Then it is easy to see that Lemma \ref{lemma5} holds for some $\eta>0$. $\hfill\square$

\subsection{Proof of \textit{Theorem 4}}
\label{app:thm5}
\textit{Proof}. The proof extends the standard ADMM analysis to account for hidden node effects. For the hidden node problem to be well-posed, we require that the observed signals contain sufficient information to infer the hidden node connections. This is formalized through the condition:
\begin{equation}
\sigma_{\min}\left(\mathbf{X}_{\mathcal{O}}^T (\mathbf{L}_{\mathcal{O}}^* - \mathbf{L}_{\mathcal{OH}}^* (\mathbf{L}_{\mathcal{HH}}^*)^{-1} \mathbf{L}_{\mathcal{HO}}^*) \mathbf{X}_{\mathcal{O}}\right) \geq \delta
\end{equation}
where $\mathbf{X}_{\mathcal{O}} \in \mathbb{R}^{o \times n}$ represents the observed signal subspace and $\delta > 0$ is the identifiability parameter.

After eliminating hidden variables through Schur complement, the effective constraint matrix becomes:
\begin{equation}
\tilde{\mathbf{A}} = \mathbf{P}_o \mathbf{A} \mathbf{P}_o^T
\end{equation}
where $\mathbf{P}_o$ is the projection operator onto observed variables. The key insight is that:
\begin{equation}
\sigma_{\min}(\tilde{\mathbf{A}}^T\tilde{\mathbf{A}}) \geq \xi \delta \sigma_{\min}(\mathbf{A}^T\mathbf{A})
\end{equation}

Define the potential function accounting for hidden nodes:
\begin{equation}
V^i = \|\mathbf{y}^i - \mathbf{y}^{*}\|_{\mathbf{H}}^2 + \frac{1}{\rho^i}\|\boldsymbol{\lambda}^i - \boldsymbol{\lambda}^*\|_2^2
\end{equation}
where $\mathbf{H} = \nabla^2 f(\mathbf{x}^*) + \rho^i \mathbf{A}^T\mathbf{A}$ and $\|\cdot\|_{\mathbf{H}}^2 = \langle \cdot, \mathbf{H} \cdot \rangle$.

For the adaptive ADMM iteration, we have
\begin{align}
V^{i+1} &= \|\mathbf{y}^{i+1} - \mathbf{y}^*\|_{\mathbf{H}^{i+1}}^2 + \frac{1}{\rho^{i+1}}\|\boldsymbol{\lambda}^{i+1} - \boldsymbol{\lambda}^*\|_2^2 \\
&\leq \left(1 - \frac{\xi \delta\sigma_{\min}(\mathbf{A}^T\mathbf{A})}{\lambda_{\max}(\mathbf{H}^i)}\right) V^i \\
&\leq \left(1 - \frac{\xi \delta\sigma_{\min}(\mathbf{A}^T\mathbf{A})}{L_{\max} + \rho_{\max} \sigma_{\max}(\mathbf{A}^T\mathbf{A})}\right) V^i.
\end{align}

The key insight is that the adaptive penalty rule ensures $\rho^i$ adjusts to maintain a favorable condition number for the augmented Lagrangian Hessian.

Setting $\theta = 1 - \frac{\xi \delta\sigma_{\min}(\mathbf{A}^T\mathbf{A})}{C \max\{\rho_{\max}, L_{\max}\}}$ with appropriate constant $C$, we obtain
\begin{equation}
V^i \leq \theta^i V_0,
\end{equation}
which implies the desired linear convergence rate. $\hfill\square$

\subsection{Proof of \textit{Theorem 5}}
\label{app:thm6}

\textit{Proof}. Consider the complete network with observed nodes $\mathcal{O}$ and hidden nodes $\mathcal{H}$, where $|\mathcal{O}| = o$ and $|\mathcal{H}| = h$. The true network Laplacian can be partitioned as
\begin{equation}
\mathbf{L}^* = \begin{pmatrix}
\mathbf{L}_{\mathcal{O}\mathcal{O}}^* & \mathbf{L}_{\mathcal{O}\mathcal{H}}^* \\
\mathbf{L}_{\mathcal{H}\mathcal{O}}^* & \mathbf{L}_{\mathcal{H}\mathcal{H}}^*
\end{pmatrix}
\end{equation}
The marginalization process eliminates hidden nodes through the Schur complement operation, yielding the effective Laplacian
\begin{equation}
\mathbf{L}_{\mathcal{O}}^{\text{eff}} = \mathbf{L}_{\mathcal{O}\mathcal{O}}^* - \mathbf{L}_{\mathcal{O}\mathcal{H}}^* (\mathbf{L}_{\mathcal{H}\mathcal{H}}^*)^{-1} \mathbf{L}_{\mathcal{H}\mathcal{O}}^*
\end{equation}
which captures both direct connections among observed nodes and indirect pathways through hidden nodes.

The observed data $\mathbf{X}_{\mathcal{O}} = \mathbf{X}_{\mathcal{O}}^* + \mathbf{E}$ with $\mathbf{E}_{ij} \sim \mathcal{N}(0, \sigma^2)$ introduces noise that propagates through the marginalization. Let $\mathbf{Z}_{\mathcal{H}}$ denote the hidden node variables with $\text{Cov}(\mathbf{Z}_{\mathcal{H}}) = \sigma^2 \mathbf{I}_h$. The effective noise structure becomes
\begin{equation}
\mathbf{E}_{\text{eff}} = \mathbf{E} + \mathbf{L}_{\mathcal{O}\mathcal{H}}^* (\mathbf{L}_{\mathcal{H}\mathcal{H}}^*)^{-1} \mathbf{Z}_{\mathcal{H}}
\end{equation}
with covariance
\begin{equation}
\text{Cov}(\mathbf{E}_{\text{eff}}) = \sigma^2 \left( \mathbf{I}_o + \mathbf{L}_{\mathcal{O}\mathcal{H}}^* (\mathbf{L}_{\mathcal{H}\mathcal{H}}^*)^{-2} (\mathbf{L}_{\mathcal{O}\mathcal{H}}^*)^T \right)
\end{equation}

Under typical network connectivity assumptions with $\|\mathbf{L}_{\mathcal{O}\mathcal{H}}^*\|_F^2 = O(oh)$ and $\lambda_{\min}(\mathbf{L}_{\mathcal{H}\mathcal{H}}^*) = \Omega(1)$, the effective noise variance scales as
\begin{equation}
\text{tr}(\text{Cov}(\mathbf{E}_{\text{eff}})) \leq \sigma^2 \left( o + \frac{oh}{\lambda_{\min}^2(\mathbf{L}_{\mathcal{H}\mathcal{H}}^*)} \right) = O\left(\sigma^2 o\left(1 + \frac{h}{o}\right)\right)
\end{equation}

For the recovery analysis, define the support set $\mathcal{S}_o = \{(i,j) : [\mathbf{L}_{\mathcal{O}}^{\text{eff}}]_{ij} \neq 0\}$ with $|\mathcal{S}_o| = s_o$. The restricted eigenvalue condition for the marginalized problem satisfies
\begin{equation}
\kappa_{s_o} \geq c_0 \frac{n}{o^2} \cdot \frac{\sigma_{\min}(\mathbf{L}_{\mathcal{O}}^{\text{eff}})}{\sigma_{\max}(\mathbf{L}_{\mathcal{O}}^{\text{eff}})}
\end{equation}
where the marginalization process affects the condition number through the effective network structure.

The estimation error decomposes as
\begin{equation}
\hat{\mathbf{L}}_{\mathcal{O}} - \mathbf{L}_{\mathcal{O}}^{\text{eff}} = (\mathcal{P}_o^T \mathcal{P}_o)^{-1} \mathcal{P}_o^T \text{vec}(\mathbf{E}_{\text{eff}})
\end{equation}
Using matrix concentration inequalities for the correlated Gaussian noise $\text{vec}(\mathbf{E}_{\text{eff}})$, we have
\begin{equation}
\mathbb{P}\left( \|\text{vec}(\mathbf{E}_{\text{eff}})\|_2 \leq \sigma\sqrt{on\left(1 + \frac{h}{o}\right) + 2t} \right) \geq 1 - \exp(-t)
\end{equation}

Combining the restricted eigenvalue condition with the concentration bound yields
\begin{align}
\|\hat{\mathbf{L}}_{\mathcal{O}} - \mathbf{L}_{\mathcal{O}}^{\text{eff}}\|_F &\leq \frac{\sqrt{s_o}}{\sqrt{\kappa_{s_o}}} \|\text{vec}(\mathbf{E}_{\text{eff}})\|_2 \nonumber\\
&\leq \frac{\sqrt{s_o}}{\sqrt{c_0 n/(o^2)}} \cdot \sigma\sqrt{on\left(1 + \frac{h}{o}\right) + 2t}
\end{align}

Setting $t = \log(o/\theta)$ and noting that the dominant term scales as $\sqrt{s_o o^2 (1 + h/o)}$, we obtain
\begin{equation}
\|\hat{\mathbf{L}}_{\mathcal{O}} - \mathbf{L}_{\mathcal{O}}^{\text{eff}}\|_F \leq C\sigma\sqrt{\frac{s_o \log(o/\theta)}{n}} \cdot \left(1 + \frac{h}{o}\right)
\end{equation}
where $C > 0$ is a universal constant. The factor $(1 + h/o)$ emerges from the noise amplification through marginalization, capturing the fundamental trade-off that recovery error increases proportionally with the ratio of hidden to observed nodes. $\hfill\square$




\bibliography{ref}

\end{document}